\documentclass[10pt,twocolumn,letterpaper]{article}

\usepackage{cvpr}              
\definecolor{cvprblue}{rgb}{0.21,0.49,0.74}
\usepackage[pagebackref,breaklinks,colorlinks,allcolors=cvprblue]{hyperref}

\usepackage{float}

\usepackage{amsmath}
\usepackage{algorithm}
\usepackage{algpseudocode}

\algnewcommand\Input{\item[\textbf{Input:}]}%
\algnewcommand\Output{\item[\textbf{Output:}]}%

\def\paperID{4177} 
\def\confName{CVPR}
\def\confYear{2026}

\title{LiSTAR: Ray-Centric World Models for 4D LiDAR Sequences in Autonomous Driving}

\author{Pei Liu$^1$\thanks{Equal contribution.} \thanks{Work done during an internship at Li Auto.}, Songtao Wang$^2$\footnotemark[1], Lang Zhang$^2$\footnotemark[1] \thanks{Project leader.}, Xingyue Peng$^2$, Yuandong Lyu$^2$, Jiaxin Deng$^2$, Songxin Lu$^2$, \\Weiliang Ma$^2$, Xueyang Zhang$^2$, Yifei Zhan$^2$, XianPeng Lang$^2$, Jun Ma\textsuperscript{\rm 1,3}\thanks{Corresponding author.}\\
\textsuperscript{\rm 1}The Hong Kong University of Science and Technology (Guangzhou)\\
\textsuperscript{\rm 2}Li Auto Inc. \\
\textsuperscript{\rm 3}The Hong Kong University of Science and Technology\\
pliu061@connect.hkust-gz.edu.cn, jun.ma@ust.hk\\
}

\begin{document}


\twocolumn[{
\renewcommand\twocolumn[1][]{#1}%
\maketitle
\begin{center}
    \centering
    \includegraphics[width=1.0\textwidth]{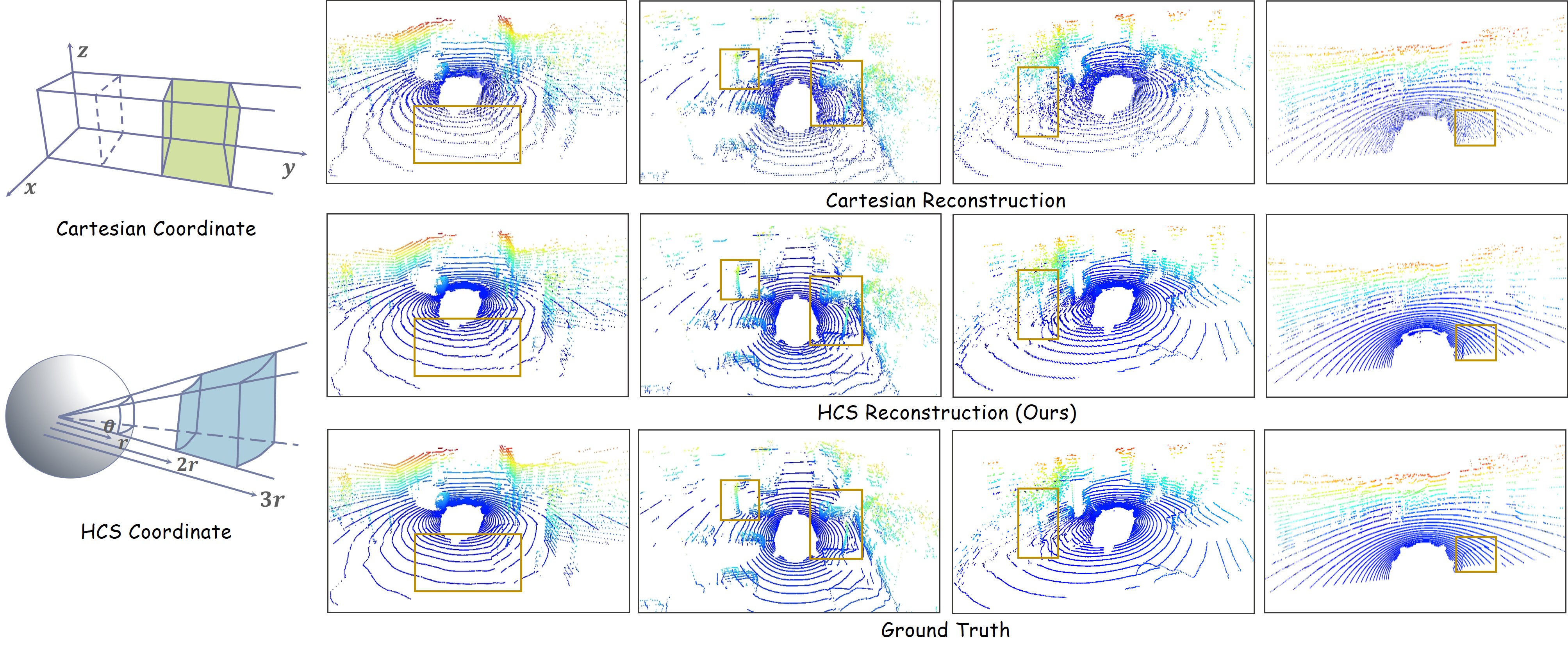}
    \captionof{figure}{Cartesian vs. HCS coordinate for LiDAR scene representation. Cartesian coordinate partitions space into uniform, axis‑aligned cubes, ignoring the native ray geometry of LiDAR. HCS coordinates divides space into angular–radial cells centered at the sensor origin, aligning with LiDAR’s ray-based sampling pattern and preserving range-dependent resolution.}
    \label{fig:fig_pre}
\end{center}
}]
\begingroup 
\renewcommand\thefootnote{*}\footnotetext{Equal contribution.}
\renewcommand\thefootnote{\ddag}\footnotetext{Project leader.}
\renewcommand\thefootnote{\S}\footnotetext{Corresponding author.}
\renewcommand\thefootnote{\dag}\footnotetext{Work done during an internship at Li Auto Inc.}
\endgroup

\begin{abstract}

   Synthesizing high-fidelity and controllable 4D LiDAR data is crucial for creating scalable simulation environments for autonomous driving. This task is inherently challenging due to the sensor's unique spherical geometry, the temporal sparsity of point clouds, and the complexity of dynamic scenes. To address these challenges, we present LiSTAR, a novel generative world model that operates directly on the sensor's native geometry. LiSTAR introduces a Hybrid-Cylindrical-Spherical (HCS) representation to preserve data fidelity by mitigating quantization artifacts common in Cartesian grids. To capture complex dynamics from sparse temporal data, it utilizes a Spatio-Temporal Attention with Ray-Centric Transformer (START) that explicitly models feature evolution along individual sensor rays for robust temporal coherence. Furthermore, for controllable synthesis, we propose a novel 4D point cloud-aligned voxel layout for conditioning and a corresponding discrete Masked Generative START (MaskSTART) framework, which learns a compact, tokenized representation of the scene, enabling efficient, high-resolution, and layout-guided compositional generation. Comprehensive experiments validate LiSTAR's state-of-the-art performance across 4D LiDAR reconstruction, prediction, and conditional generation, with substantial quantitative gains: reducing generation MMD by a massive 76\%, improving reconstruction IoU by 32\%, and lowering prediction L1 Med by 50\%. This level of performance provides a powerful new foundation for creating realistic and controllable autonomous systems simulations. Project link: \href{https://ocean-luna.github.io/LiSTAR.gitub.io/}{LiSTAR}.

\end{abstract}    
\section{Introduction}
\label{sec:intro}
World models, which aim to internalize environmental dynamics by learning generative predictors, have demonstrated strong capabilities across a wide range of visual and interactive tasks and are now increasingly explored for autonomous driving \cite{hu2023gaia, mei2024dreamforge}. Recent progress has largely focused on structured modalities like videos and occupancy grids, whose dense organization fits well with established processing pipelines \cite{gao2023magicdrive, wang2024occsora}. By contrast, LiDAR remains understudied despite its importance for accurate 3D geometry and all-weather perception. The sparse, unordered, and irregular nature of LiDAR point clouds \cite{kong2023robo3d, liu2023segment, xu20244d} poses fundamental challenges for generative modeling, limiting the direct adoption of techniques designed for regularly structured data.

Despite recent progress in LiDAR scene synthesis \cite{zyrianov2022learning, li2023pointdmm, yu2023video}, significant hurdles remain. A primary challenge stems from conventional voxelization, which converts LiDAR returns into dense Cartesian grids. As illustrated in Fig. \ref{fig:fig_pre}, this approach overlooks the native ray-based sampling geometry of spinning sensors, leading to quantization artifacts and distorted structural patterns that impair fidelity \cite{zhang2023copilot4d, liang2025lidarcrafter, li2025uniscene}. Furthermore, the inherent sparsity and non-uniform sampling of point clouds complicate the preservation of temporal coherence, often resulting in flickering surfaces or inconsistent dynamic object alignment \cite{huang2024vbench}. Finally, for controllable synthesis, the prevalent reliance on temporal Bird's-Eye-View (BEV) layouts \cite{nakashima2024lidar, nakashima2025fast} as conditional inputs imposes a critical bottleneck. This 2D projection inherently flattens the rich 3D world, constraining the ability to precisely guide generation or manipulate objects in full 3D space, a capability crucial for targeted scenario design and safety evaluation.

To address these challenges, we introduce LiSTAR, a novel world model built upon a pioneering Hybrid-Cylindrical-Spherical (HCS)-based 4D Vector Quantised-Variational AutoEncoder (VQ-VAE) \cite{van2017neural,caccia2019deep} to learn a discrete representation of LiDAR scenes. LiSTAR begins with a novel HCS representation, the first of its kind for LiDAR world models, which aligns with the sensor's native scanning geometry to preserve ray structure and mitigate distortions. Building on this representation, our Spatio-Temporal Attention with Ray-Centric Transformer (START) module explicitly models feature correlations along sensor rays across time, enforcing robust spatial and temporal consistency. Finally, to enable controllable synthesis, we introduce a novel 4D point cloud-aligned voxel layout as a conditioning mechanism. A discrete Masked Generative START (MaskSTART) pipeline then operates on the learned VQ tokens, conditioned on these layouts, to achieve efficient, high-fidelity generation. Collectively, these synergistic innovations enable the creation of 4D LiDAR scenes that are not only physically faithful but also precisely controllable, paving the way for more realistic and targeted autonomous driving simulation.

We conduct extensive experiments on the large-scale nuScenes benchmark, evaluating LiSTAR on a suite of tasks including point cloud reconstruction, prediction, and generation. In both unconditional and layout-conditioned settings, LiSTAR consistently outperforms state-of-the-art baselines. Beyond quantitative metrics, we demonstrate that the framework's ability to produce controllable, temporally consistent LiDAR sequences unlocks novel downstream applications. 
The main contributions of this work are:

\begin{itemize}
\item We present LiSTAR, a 4D LiDAR world model that unifies HCS representation, START, and MaskSTART into a single end‑to‑end framework, explicitly tailored to LiDAR’s acquisition geometry and temporal dynamics for world models of autonomous driving.

\item We propose an HCS coordinate voxelization scheme that preserves the native ray structure and range resolution, effectively mitigating geometric distortion caused by conventional Cartesian discretization.

\item We design the START module, which models feature correlations along LiDAR rays to capture spatial structure and temporal dependencies jointly, ensuring geometric fidelity and frame‑to‑frame consistency.

\item We introduce a MaskSTART pipeline for 4D LiDAR sequences that supports fine-grained semantic conditioning on 4D point cloud-aligned voxel layout. This approach enables controllable and diverse scenario synthesis, allowing for precise manipulation and generation of complex scene structures.
 
\item We achieve state‑of‑the‑art performance on a large‑scale autonomous driving benchmark for both point cloud reconstruction, prediction, and generation, and demonstrate LiSTAR’s utility in realistic, controllable simulation scenarios.
\end{itemize}

\section{Related Work}
\label{sec:relatedwork}

\subsection{3D Representation for Point Clouds}  
Choosing an effective 3D representation is critical for point cloud generation.  
Point-based approaches, such as PointNet and PointNet++~\cite{lin2017structured,qi2017pointnet,qi2017pointnet++}, directly operate on raw points, aggregating local and global features to encode spatial context.  
Voxelization~\cite{qi2016volumetric,yang2018pixor,liu2021pvnas} discretizes the space into dense grids but is memory-intensive, leading to sparse convolution designs~\cite{choy20163d,graham20183d} that skip empty cells.  
Projection-based representations are also popular: BEV~\cite{lang2019pointpillars,yang2018pixor} vertically projects points onto a planar map, while range images~\cite{milioto2019rangenet++,zyrianov2022learning,ran2024towards,nakashima2024lidar,hu2024rangeldm} map points to polar coordinates to form 2.5D grids.  
Recent works leverage VAE-family models~\cite{kingma2013auto,van2017neural,esser2021taming} for latent compression, e.g., VQ-VAE~\cite{van2017neural,caccia2019deep} learns discrete codebooks for compact feature tokens.  
For simulation, ray-casting pipelines~\cite{amini2022vista,dosovitskiy2017carla,manivasagam2020lidarsim,nakashima2021learning} reproduce ray-drop patterns from virtual assets.  
Finally, implicit neural representations, such as NeRF~\cite{mildenhall2021nerf}, allow differentiable rendering of point clouds from learned occupancy or semantic fields~\cite{li2025uniscene,zhang2024nerf}.

\subsection{World Models for Point Clouds}  
World models predict future observations from historical states and agent actions, enabling agents to model temporal dynamics.  
While early work focused on image/video prediction~\cite{hu2023gaia,russell2025gaia,zhao2025drivedreamer,zhao2025drivedreamer4d,wu2024drivescape}, recent studies extend to structured 3D data.  
In 3D occupancy world models~\cite{zheng2024occworld,wei2024occllama,zuo2025gaussianworld,li2025occmamba,gu2024dome}, discrete volumetric tokens are predicted to maintain spatial consistency.  
Point cloud world models~\cite{zhang2023copilot4d,zyrianov2025lidardm,yang2024visual} predict temporal LiDAR sequences by combining latent tokenization and generative backbones.  
For instance, Copilot4D~\cite{zhang2023copilot4d} encodes LiDAR frames with VQ-VAE and applies discrete diffusion for forecasting, while LiDAR-DM~\cite{zyrianov2025lidardm} adapts diffusion transformers for long-horizon predictions.  
These approaches highlight the promise and current limitations of scalable, geometry-aware token-based generative modeling for 4D LiDAR.

\subsection{Diffusion Models}  
Diffusion models learn a forward process that progressively corrupts data with noise, and a reverse process to recover the original signal.  
For continuous data, Gaussian perturbations~\cite{ho2020denoising,song2020denoising,song2020score,rombach2022high} are widely used due to favorable statistical properties enabling stable training objectives.  
MaskGIT~\cite{chang2022maskgit} replaces Gaussian noise with aggressive token masking and BERT-style training~\cite{devlin2019bert}, outperforming Gaussian diffusion in several domains, including video~\cite{yu2023magvit} and point clouds~\cite{xiong2023learning}.  
Beyond diffusion, flow matching ~\cite{lipman2022flow} learns continuous-time flows between base and target distributions, allowing ODE-based sampling with convergence guarantees, and has been applied to sequences~\cite{geng2025mean} and spatial modalities.  
Recent works adapt these generative paradigms to 3D point clouds, using token-based pipelines, view-consistent constraints, and geometry-aware noise schedules to improve spatial fidelity and temporal stability, making them promising backbones for LiDAR world modeling.


\section{Methodology}

We introduce LiSTAR, a novel generative world model for 4D LiDAR synthesis, composed of two synergistic components: an HCS-based 4D VQ-VAE for representation learning and a MaskSTART model for prediction and generation.

\begin{figure*}
    \centering
    \includegraphics[width=1.0\textwidth]{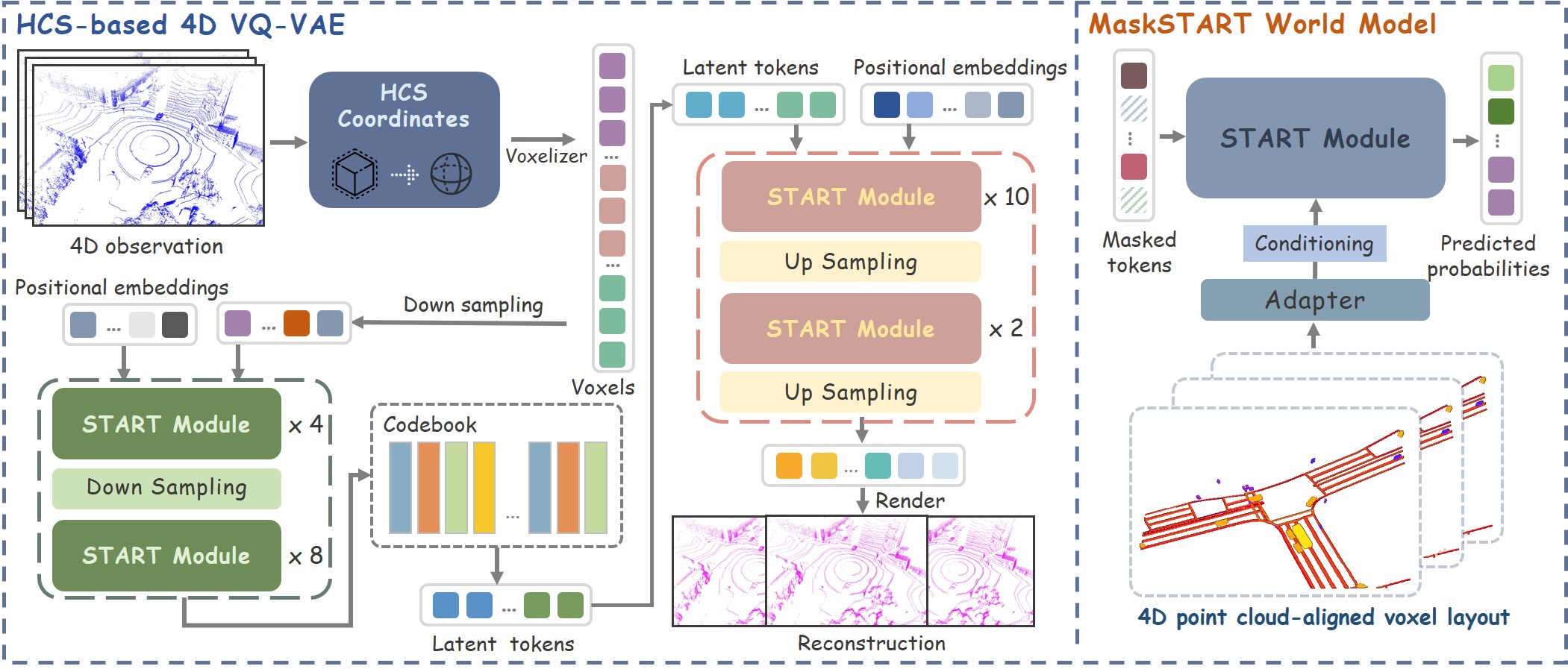}
    \caption{Illustration of the LiSTAR framework for 4D LiDAR sequence reconstruction and generation. The framework begins by voxelizing LiDAR point clouds into a spherical coordinate representation, which is downsampled and processed by multiple START modules in the encoder to extract semantic-rich latent tokens. The decoder reconstructs detailed 4D sequences by up-sampling tokens with additional START modules. The MaskSTART component facilitates controllable and diverse generation by predicting masked tokens using a bidirectional transformer, conditioned on 4D point cloud-aligned voxel layouts. This design captures spatiotemporal dependencies while preserving fine-grained geometric details.} 
    \label{fig:figure_vqvae}
\end{figure*}

The HCS-based 4D VQ-VAE, shown in Fig. \ref{fig:figure_vqvae} (left), first transforms the input LiDAR sequence into a compact, discrete latent space. The encoder employs stacked START blocks, featuring Spatial Ray-Centric Attention (SRA) and Cyclic-Shifted Temporal Causal Attention (CSTA), to capture spatio-temporal dynamics and produce a quantized codebook representation effectively. This discrete representation then serves as the foundation for the MaskSTART model (Fig. \ref{fig:figure_vqvae}, right), a unified framework that performs masked generative modeling for both prediction and conditional generation. In the generation task, it conditions on 4D point cloud-aligned voxel layouts, which are fused via a zero-initialized adapter to guide the synthesis of realistic and semantically consistent sequences. Further details on the algorithmic procedures for reconstruction, prediction, and generation are provided in the Appendix.

\subsection{HCS Coordinate Voxelization}

Conventional Cartesian voxelization forces a trade-off between fidelity and efficiency: high-resolution grids needed for detail are massively sparse and computationally expensive. We overcome this by proposing a voxelization scheme in an HCS Coordinate System that mirrors the native spherical projection of LiDAR sensors, as shown in Fig.~\ref{fig:fig_pre}. Our method partitions space into bins of constant angular resolution, preserving geometric details at all ranges while yielding a compact and efficient representation.
Formally, a point cloud is defined as:
\[
P = \{ \mathbf{p}^{(n)} \in \mathbb{R}^3 \;\mid\; 1 \leq n \leq N \},
\] 
where each point $\mathbf{p}^{(n)} = (x^{(n)}, y^{(n)}, z^{(n)})$ is expressed in Cartesian coordinate. We map these points into the HCS coordinate system $(\rho^{(n)}, \theta^{(n)}, \phi^{(n)})$ using:  
\begin{equation}
\label{eq4}
\begin{cases}
\rho^{(n)} = \sqrt{(x^{(n)})^2 + (y^{(n)})^2 }, \\
\theta^{(n)} = \arctan2(y^{(n)}, x^{(n)}), \\
\phi^{(n)} = \arctan2(z^{(n)}, \rho^{(n)}).
\end{cases}
\end{equation}
The voxelization is performed by a function $\text{V}:(\rho,\theta,\phi)\rightarrow(i,j,k)$, which quantizes a point's continuous HCS coordinates into a discrete integer tuple. This tuple $(i,j,k)$ indexes a specific bin within a 3D grid along the radial, angular, and axial dimensions. The final output is a binary occupancy grid $\mathcal{G}$, where an element $\mathcal{G}_{i,j,k}=1$ if the corresponding voxel is non-empty, and 0 otherwise. This formulation provides a structured and compact encoding of the raw point cloud, ideal for consumption by subsequent network layers.

\subsection{START Module}
\begin{figure}[!ht] 
    \centering
    \includegraphics[width=0.5\textwidth]{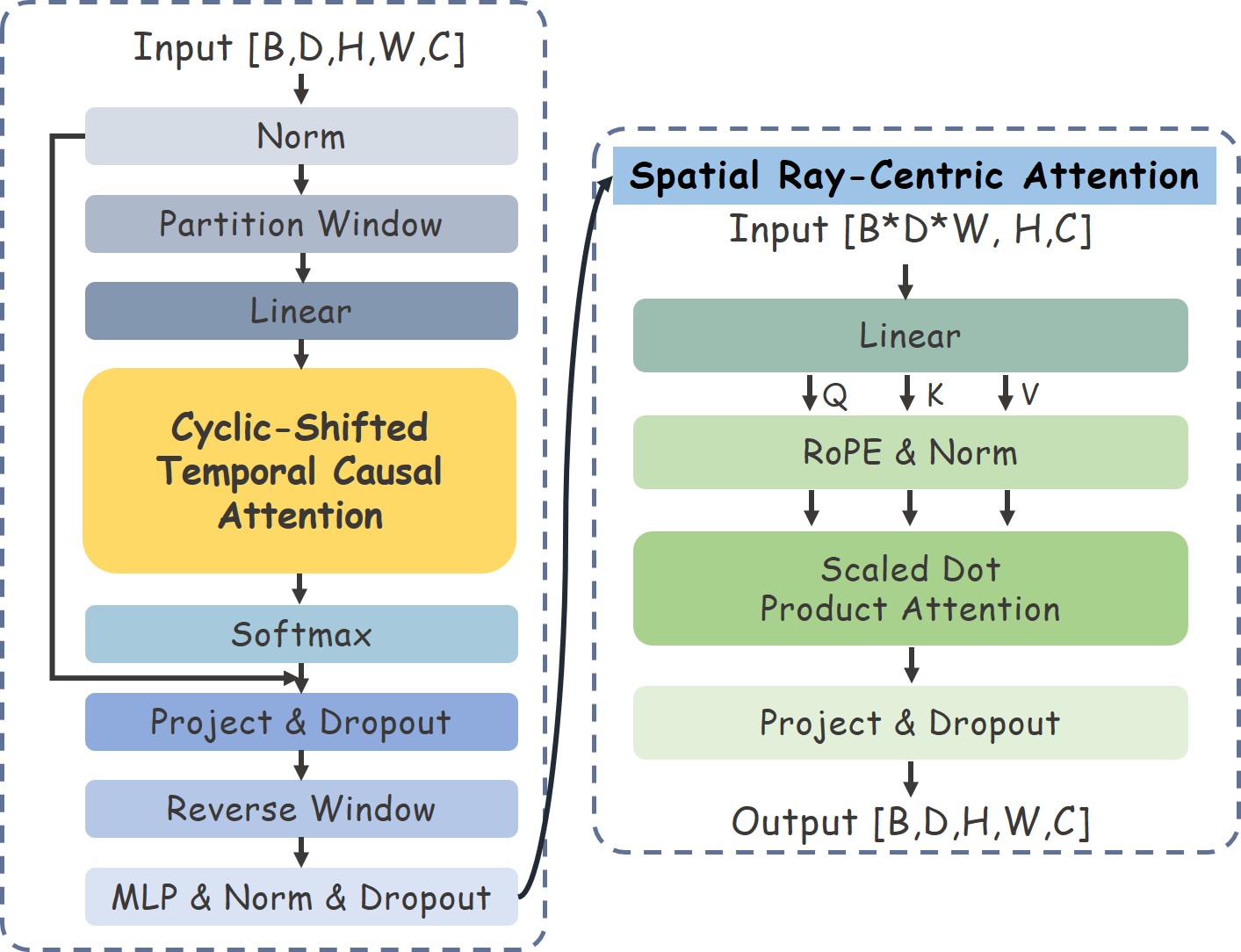}
    \caption{An illustration of our START module. It processes a 4D feature map of shape $[B, D, H, W, C]$, where $D$ is the temporal dimension. It is composed of two key components: (1) a CSTA block that operates on windowed features to efficiently model temporal dependencies, and (2) an SRA block that processes features reshaped to $[B*D, H, W, C]$ to capture spatial correlations along the ray dimension.} 
    \label{fig:figure_start}
\end{figure}

We introduce the START module, a novel 4D attention mechanism specifically designed for sequential LiDAR data. START operates in a causal temporal manner to capture motion patterns while leveraging spatial ray-centric attention, explicitly aligned with the intrinsic geometry of LiDAR sensors. This formulation enables the model to capture both spatiotemporal dependencies and fine-grained structural relationships effectively, as illustrated in Fig. \ref{fig:figure_start}. 

\subsubsection{Spatial Ray-Centric Attention} 

To explicitly encode the intrinsic ray-like structure of LiDAR scans, we introduce the Ray-Centric Attention (RA) layer. It operates on a dense tensor $V_{\rho \theta \phi}\in \mathbb{R}^{l\times h \times w}$, where $l$, $h$, and $w$ correspond to the ray, vertical, and horizontal angular dimensions, respectively. To efficiently process this representation, we first unfold ($\text{F}$) the tensor $V$ along its ray dimension $l$, transforming it into a 2D matrix of size $d\times l$, where $d=h\times w$. A standard self-attention mechanism is then applied to a normalized version of this flattened representation, denoted $V_n$. The output is computed as:
\begin{equation}
    V_n = \text{Norm}(\text{F}(V)),
\end{equation}
\begin{equation}
V' = softmax \left (\frac{V_nW_Q(V_nW_K)^T}{\sqrt{d_k}} \right )V_nW_v, 
\end{equation}
where $W_Q$, $W_K$, and $W_V$ are linear projection matrices that map the input features into query, key, and value spaces. This mechanism enables each ray to aggregate information from all other rays based on their learned similarities, effectively capturing the global context of the 3D scene.

Spatial RA (SRA) block embeds the proposed RA layer within a pre-norm Transformer architecture. The process involves a residual connection after the RA-layer, followed by Layer Normalization (LN) and a Feed-Forward Network (FFN) with a second residual connection. Formally, this is expressed as:
\begin{equation}
\label{eq:sra_block_compact}
V_{\text{SRA}} = \text{F}^{-1} \left( (V' + f(V')) + g(\text{LN}(V' + f(V'))) \right),
\end{equation}
where $f(\cdot)$ and $g(\cdot)$ represent the RA layer and the FFN, respectively. $\text{F}^{-1}$ indicates the inverse function of $\text{F}$. This residual-in-residual design ensures stable training while enhancing feature expressiveness.

Applying standard self-attention to the full 4D voxel grid is not only computationally prohibitive but also structurally agnostic to the underlying LiDAR geometry. It fails to exploit the inherent radial dependencies captured by sensor rays. Our SRA addresses this by restricting attention computations to the radial dimension. This targeted approach allows the model to efficiently reason about occlusions and spatial relationships along each line of sight. Consequently, SRA captures fine-grained local structures while maintaining global context, all within a feasible computational budget.

\subsubsection{Cyclic-Shifted Temporal Causal Attention}

To effectively model 4D LiDAR sequences, our model must address two fundamental challenges: the spatial discontinuity arising from spherical coordinate projection, and the need for causal temporal modeling. To this end, we propose the CSTA module, which is composed of two specialized mechanisms. First, Cyclic-Shifted Window Attention (CSWA) restores spatial continuity across the azimuthal seam by leveraging a shifted-windowing scheme inspired by the Swin Transformer \cite{liu2021swin}. Second, Temporal Causal Attention (TCA) enforces a strict chronological order to learn valid motion patterns without information leakage from the future.

\textbf{Cyclic-Shifted Window Attention}. The projection of LiDAR's HCS coordinates onto a discrete grid introduces a critical spatial discontinuity. Points that are physically adjacent across the azimuthal seam ($0^\circ/360^\circ$) are placed at opposite ends of the tensor's width dimension, artificially severing their geometric relationship.

This issue can be conceptualized by considering the mapping of a local neighborhood. Let $\mathcal{N}(p)$ denote a connected 3D neighborhood of a point $p$ located on the azimuthal boundary. The projection onto a tensor of width W splits this single neighborhood into two disjoint sets of indices at the extremes of the tensor dimension:
\begin{equation}
\mathcal{N}(p) \xrightarrow{\text{Projection}} { \cdots, w_{W-2}, w_{W-1} } \cup { w_0, w_1, \cdots }.
\end{equation}
Consequently, standard network operators with fixed receptive fields (e.g., convolutions, window attention) fail to process this neighborhood cohesively. They perceive the two parts as maximally distant, leading to feature artifacts and an incomplete understanding of the global scene structure.

\begin{algorithm}[H]
\caption{Cyclic-Shifted Window Attention (CSWA)}
\label{alg:cswa}
\begin{algorithmic}[1]
\Input Input feature map $X_l \in \mathbb{R}^{B \times D \times H \times W \times C}$ from layer $l$, window size $(M_D, M_H, M_W)$.
\Output Output feature map $X_{l+1}$ from the subsequent layer.

\State \Comment{\textbf{Stage 1: Standard Window MSA (W-MSA)}}
\State $X' \gets \text{LayerNorm}(X_l)$
\State $X_{\text{windows}} \gets \text{WindowPartition}(X', (M_D, M_H, M_W))$ \Comment{Partition into non-overlapping windows}
\State $A_{\text{windows}} \gets \text{MSA}(X_{\text{windows}})$ \Comment{Apply Multi-Head Self-Attention within each window}
\State $A \gets \text{WindowReverse}(A_{\text{windows}}, (D, H, W))$ \Comment{Merge windows back}
\State $X_l \gets X_l + A$ \Comment{First residual connection}
\State $X_l \gets X_l + \text{MLP}(\text{LayerNorm}(X_l))$ \Comment{Second residual connection with MLP}

\Statex
\State \Comment{\textbf{Stage 2: Shifted Window MSA (SW-MSA)}}
\State $X'' \gets \text{LayerNorm}(X_l)$
\State $X_{\text{shifted}} \gets \text{CyclicShift}(X'', (M_D/2, M_H/2, M_W/2))$ \Comment{Cyclic shift along the azimuthal (W) dimension}

\State \Comment{Generate mask to prevent attention between non-adjacent regions}
\State $M \gets \text{GenerateMask}(D, H, W, (M_D, M_H, M_W))$ 

\State $X'_{\text{windows}} \gets \text{WindowPartition}(X_{\text{shifted}}, (M_D, M_H, M_W))$
\State $A'_{\text{windows}} \gets \text{MSA}(X'_{\text{windows}}, \text{mask}=M)$ \Comment{Apply MSA with the generated mask}
\State $A'_{\text{shifted}} \gets \text{WindowReverse}(A'_{\text{windows}}, (D, H, W))$

\State $A' \gets \text{CyclicShift}(A'_{\text{shifted}}, (0, 0, M_W/2))$ \Comment{Reverse the cyclic shift}
\State $X_{l+1} \gets X_l + A'$ \Comment{Third residual connection}
\State $X_{l+1} \gets X_{l+1} + \text{MLP}(\text{LayerNorm}(X_{l+1}))$ \Comment{Fourth residual connection with MLP}

\State \Return $X_{l+1}$
\end{algorithmic}
\end{algorithm}

\begin{table*}[t]
\centering
\caption{\textbf{LiSTAR significantly outperforms the previous state-of-the-art method in point cloud reconstruction.}  Our method demonstrates significant improvements across all metrics, achieving a 32\% relative increase in IoU and a 60\% reduction in MMD, indicating superior geometric accuracy and distribution similarity. Best results are in bold. ($\uparrow$: Higher is better, $\downarrow$: Lower is better).}
\label{tab:reconstruction}
\begin{tabular}{l|cccccc}
\toprule
Method & {IoU$\uparrow$} & {Chamfer$\downarrow$} & {MMD ($10^{-4}$)$\downarrow$} & {JSD$\downarrow$} \\
\midrule
OpenDWM \cite{opendwm, chen2024unimlvg, ni2025maskgwm} & 0.441 & 0.029 & 0.152 & 0.076\\
Ours & \textbf{0.583} (32\%$\uparrow$) & \textbf{0.017} (41\%$\downarrow$)& \textbf{0.061} (60\%$\downarrow$)& \textbf{0.056} (26\%$\downarrow$)\\
\bottomrule
\end{tabular}
\end{table*}

\begin{table*}[t]
\centering
\caption{\textbf{LiSTAR sets a new state-of-the-art in point cloud prediction.} The table compares our method with previous state-of-the-art approaches on the nuScenes dataset. LiSTAR achieves a 17\% reduction in Chamfer distance and a 50\% reduction in L1 Med. Metrics shown in blue are evaluated within a ±70m ROI. Best results are in bold. $\downarrow$ indicates lower is better.
}
\label{tab:prediction}
\begin{tabular}{l|ccccc}
\toprule
\textbf{Method} & \textcolor{blue}{Chamfer}$\downarrow$ & \textcolor{blue}{L1 Med}$\downarrow$ & \textcolor{blue}{AbsRel Med}$\downarrow$ & L1 Mean$\downarrow$ & AbsRel$\downarrow$\\
\midrule
SPFNet \cite{weng2021inverting}   & 2.24 & -    & -    & 4.58 & 34.87 \\
S2Net \cite{weng2022s2net}    & 1.70 & -    & -    & 3.49 & 28.38\\
4D-Occ \cite{khurana2023point} & 1.41 & 0.26 & 4.02 & 1.40 & 10.37 \\
Copilot4D \cite{zhang2023copilot4d} & 0.36 & 0.10 & 1.30 & 1.30 & 8.58\\
\midrule
Ours & \textbf{0.30} (17\%$\downarrow$) & \textbf{0.05} (50\%$\downarrow$) & \textbf{0.96} (26\%$\downarrow$)& \textbf{0.76} (42\%$\downarrow$)& \textbf{4.92} (43\%$\downarrow$)\\
\bottomrule
\end{tabular}
\end{table*}

To address the boundary discontinuity induced by spherical coordinate unwrapping, we propose CSWA. CSWA explicitly models the periodic nature of the azimuthal dimension, enabling information flow across the artificial seam. The mechanism operates in two alternating stages. The procedure, detailed in Alg. \ref{alg:cswa}, alternates between two configurations. A standard Window MSA (W-MSA) first computes self-attention within local, non-overlapping windows for efficient feature extraction. Subsequently, a Shifted Window MSA (SW-MSA) block introduces a cyclic shift along the azimuthal dimension. This realigns the window grid, enabling cross-window connections, particularly across the $0^\circ/360^\circ$ seam. A carefully designed attention mask ensures that interactions are confined to valid local regions in the shifted configuration before the shift is reversed.

By alternating these standard and shifted window configurations, CSWA achieves a global receptive field with linear complexity, efficiently restoring the topological continuity of the spherical space.

\textbf{Temporal Causal Attention}. 
Beyond static spatial features, modeling temporal dynamics is crucial for interpreting motion and ensuring coherence across LiDAR frames. Standard attention mechanisms are permutation-invariant and thus non-causal, allowing information to leak from future frames, which is invalid for predictive tasks.

To address this, we introduce TCA, a mechanism designed to model scene evolution while strictly adhering to the arrow of time. TCA extends the causal constraint to a history of 
$L$ preceding frames, $\{X_{t-L}, X_{t-L+1}, \cdots, X_{t-1}\}$, allowing the model to capture long-range temporal dependencies.
Queries $Q_t$ are generated from the current frame $X_t$, while a unified set of keys $K_{hist}$ and values $V_{hist}$ is created by concatenating the respective projections from all $L$ past frames. The attention mechanism then aggregates information from the entire history as follows:

\begin{equation}
\text{TCA}(X_t) = \text{softmax} \left( \frac{Q_t K_{\text{hist}}^T}{\sqrt{d_k}} \right) V_{\text{hist}}.
\end{equation}
This formulation ensures that the model's output for time t depends on past and present information, enabling it to learn robust and complex motion patterns from a rich temporal context.

We integrate TCA by interleaving it with CSWA layers. This layered structure allows the model to jointly refine spatial details and update temporal states, creating a comprehensive 4D representation. This unified approach is essential for tasks requiring both spatial integrity and temporal coherence, such as dynamic scene reconstruction and motion forecasting.


\begin{table*}[t]
\centering
\caption{\textbf{LiSTAR demonstrates superior performance in LiDAR generation.} Our method significantly outperforms the OpenDWM baseline in both geometric accuracy and distributional fidelity. As shown, LiSTAR reduces the MMD by 76\% and cuts the Chamfer distance by over 50\% across different evaluation ranges: 30m (magenta), 40m (green), and 70m (blue). This highlights our model's ability to generate significantly more realistic and accurate point cloud sequences. Best results are in bold.}
\label{tab:generation}
\begin{tabular}{l|ccccc}
\toprule
\textbf{Method} &  \textcolor{magenta}{Chamfer}$\downarrow$ &  \textcolor{green}{Chamfer}$\downarrow$ &  \textcolor{blue}{Chamfer}$\downarrow$ &MMD ($10^{-4}$)$\downarrow$ & \textbf{JSD}$\downarrow$ \\
\midrule

OpenDWM & 1.88 & 2.57 & 3.35 & 41.14 & 0.31\\

Ours & \textbf{0.72} (62\%$\downarrow$)& \textbf{1.21} (53\%$\downarrow$)& \textbf{1.53} (54\%$\downarrow$)& \textbf{9.94} (76\%$\downarrow$)& \textbf{0.30}\\
\bottomrule
\end{tabular}
\end{table*}

\begin{table*}
\centering
\caption{\textbf{Our proposed HCS coordinate achieves superior performance.} The table presents a direct comparison against standard Cartesian and Polar coordinates, where our method demonstrates significant gains across all metrics. For example, it boosts IoU by 16\% over the next-best polar representation. Bold denotes the best performance.}
\label{tab:ab2}
\begin{tabular}{ccc|cccccc}
\toprule
\textbf{Cartesian} & \textbf{Polar} & \textbf{HCS} & {IoU$\uparrow$} & {Chamfer$\downarrow$} & {MMD ($10^{-4}$)$\downarrow$} & {JSD$\downarrow$} \\
\midrule
\checkmark &  &  & 0.414 & 0.039 & 0.475 & 0.086\\
&\checkmark &  & 0.476 & 0.023 & 0.072 & 0.067 \\
&&\checkmark &  \textbf{0.554} (16\%$\uparrow$) & \textbf{0.020} (13\%$\downarrow$)& \textbf{0.065} (10\%$\downarrow$)& \textbf{0.060} (10\%$\downarrow$)\\
\bottomrule
\end{tabular}
\end{table*}

\section{Experiments}
\label{experiments}
In this section, we aim to conduct experiments to investigate the following questions: (1) Does LiSTAR achieve state-of-the-art performance on the autonomous driving benchmark across point cloud reconstruction, forecasting, and generation tasks? (2) How crucial are the respective contributions of the HCS representation and the START module? (3) Qualitatively, how effective is LiSTAR at generating temporally coherent and high-fidelity reconstructions?

\subsection{Datasets and Experiment Setting} 
Our experiments are conducted on the large-scale nuScenes dataset \cite{caesar2020nuscenes}, which provides dense, 360-degree point clouds from a 128-beam LiDAR. We utilize the official train/val split, focusing on keyframes from diverse urban scenarios. To create a fixed-size input, each raw point cloud is downsampled to 2048 points via Farthest Point Sampling (FPS). For the prediction task, we define the operational range as [-70, 70]m in x/y and [-4.5, 4.5]m in z, while for the generation task, the range is [-50, 50]m in x/y and [-3, 5]m in z. All models are trained for 60k steps on 64 H20 GPUs using the AdamW optimizer with a learning rate of 5e-5. We use a per-GPU batch size of 2 and train with bf16 precision for computational efficiency.

\subsection{Metrics} 
To comprehensively evaluate our method, LiSTAR, across point cloud reconstruction, prediction, and generation, we assess both per-sample geometric fidelity and overall distributional similarity. For geometric fidelity, we measure volumetric accuracy using Intersection over Union (IoU) and quantify point-wise discrepancies using Chamfer distance, L1 distance (both Mean and Median), and Absolute Relative error (AbsRel). Notably, for the LiDAR generation task, Chamfer distance is evaluated across multiple ranges (30m, 40m, and 70m) to assess fidelity at varying distances. To evaluate the distributional quality of the generated set, we employ Maximum Mean Discrepancy (MMD) and Jensen-Shannon Divergence (JSD), which are critical for judging the realism and diversity of the generated point cloud sequences.

\subsection{Reconstruction Results}
As shown in Table \ref{tab:reconstruction}, LiSTAR significantly outperforms the previous state-of-the-art method, OpenDWM, in the task of point cloud reconstruction. Our method achieves substantial gains across the board, with a 26\% relative increase in IoU and a 57\% reduction in MMD. These results demonstrate that LiSTAR not only reconstructs scene geometry more accurately by achieving a higher IoU and lower Chamfer Distance, but also captures the data distribution with much higher fidelity, reflected in lower MMD and JSD. This comprehensive improvement validates the effectiveness of our proposed architecture for high-quality reconstruction.

\begin{figure}
    \centering
    \includegraphics[width=1.0\linewidth]{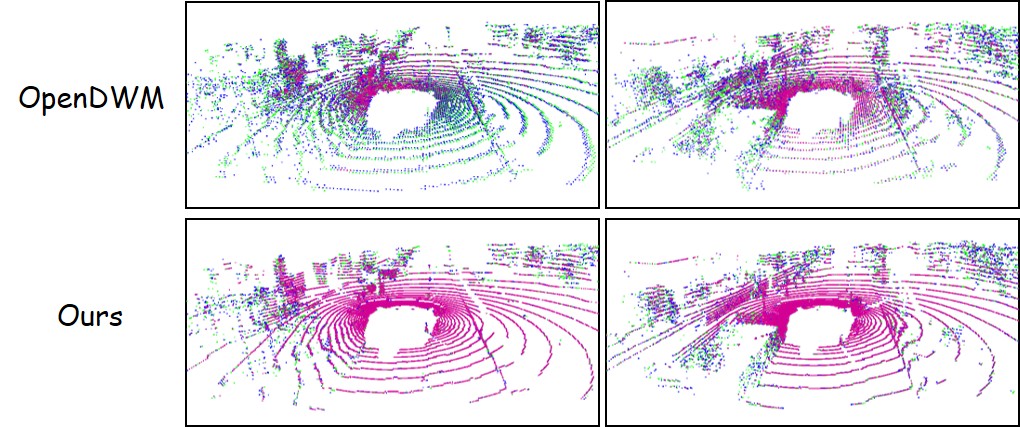}
    \caption{Qualitative comparison of point cloud reconstruction. The visualization overlays predictions with the ground truth: magenta (correct intersection), green (missed ground truth), and blue (artifacts). Our method consistently yields more complete reconstructions (denser magenta) with significantly fewer artifacts (less blue), demonstrating superior accuracy.}
    \label{fig:rec_result}
\end{figure}

\begin{figure*}
    \centering
    \includegraphics[width=1.0\linewidth]{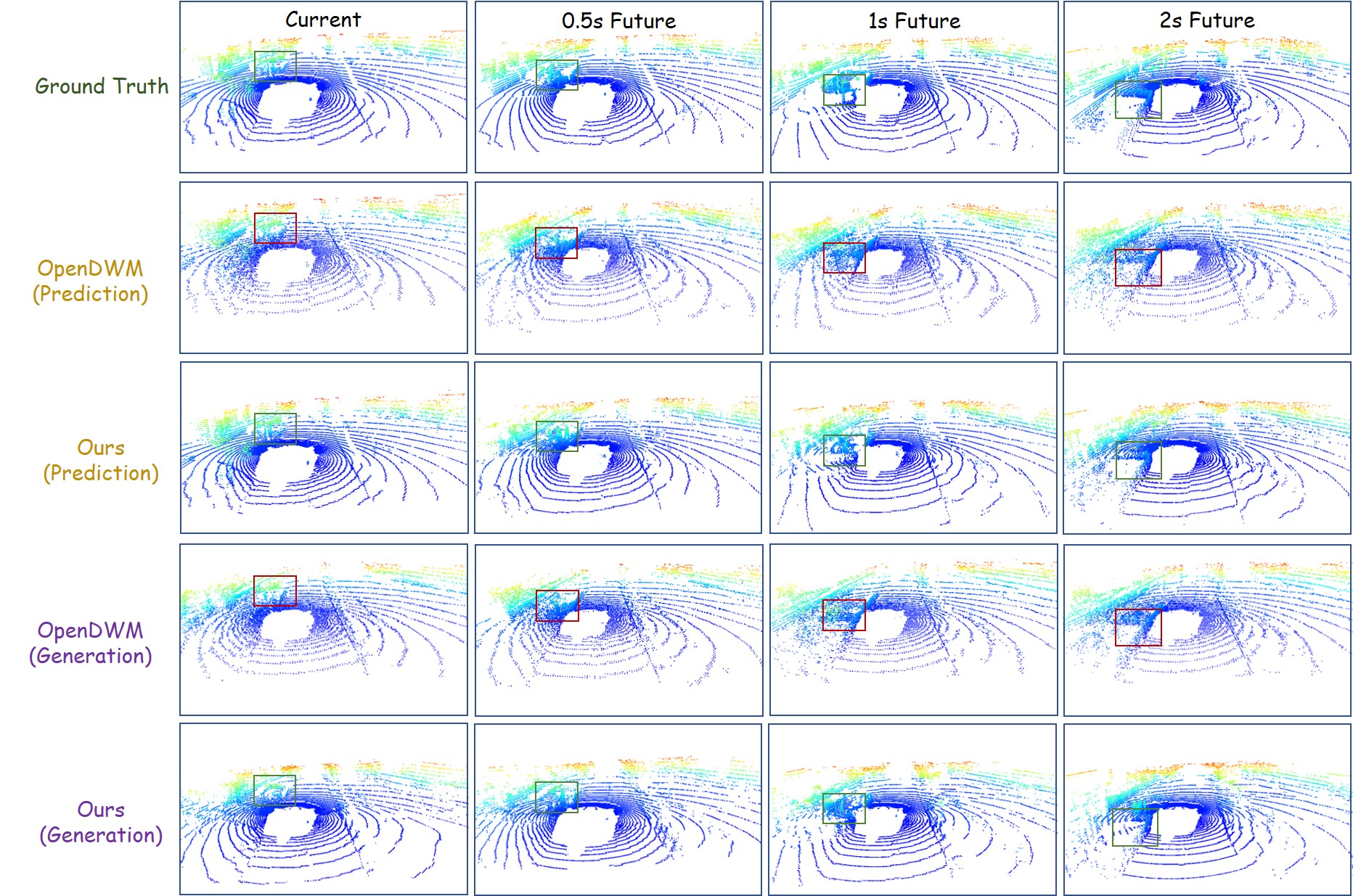}
    \caption{Qualitative results for prediction and generation. We compare our method with OpenDWM against the ground truth for future horizons up to 2s. Our method consistently produces sharper and more accurate results for both static background and dynamic objects (highlighted) compared to the baseline. The baseline's predictions and generations degrade significantly over time, losing structural detail.}
    \label{fig:pre_gen}
\end{figure*}

\subsection{Prediction Results}
The results for 1\,s future prediction are presented in Table \ref{tab:prediction}. Our method, LiSTAR, significantly outperforms all baselines, establishing a new state-of-the-art. Notably, for the 1s future horizon, LiSTAR reduces the Chamfer distance by 17\% and the L1 Med by a remarkable 50\% compared to Copilot4D. This comprehensive improvement at the 1\,s horizon validates the effectiveness of our architecture in producing highly accurate and reliable future predictions.

\subsection{Generation Results}

For the LiDAR generation task, we compare LiSTAR with the OpenDWM baseline, presenting quantitative results in Table \ref{tab:generation}. Our method demonstrates a substantial advantage across all metrics. Most notably, LiSTAR achieves a remarkable 76\% reduction in Maximum Mean Discrepancy (MMD), indicating that the distribution of our generated point clouds is significantly closer to the ground truth. In terms of geometric accuracy, our model consistently reduces the Chamfer distance by over 50\% across all evaluated ranges: 30m (magenta), 40m (green), and 70m (blue). These results collectively validate the superior capability of our model in generating high-fidelity and physically realistic LiDAR sequences.

\subsection{Ablation Study}
\subsubsection{Analysis of Coordinate Representation}
Table~\ref{tab:ab2} presents our ablation study on coordinate systems, demonstrating the clear superiority of our proposed HCS representation. HCS substantially outperforms both Cartesian and the stronger Polar coordinate baselines across all metrics. Specifically, it achieves an IoU of 0.554, marking a significant 16\% relative improvement over Polar coordinates. This result strongly validates the advantage of HCS in providing a more powerful representation for LiDAR data.

\subsubsection{Effectiveness of the START Module}
Table~\ref{tab:ab3} validates the synergistic design of our START module, demonstrating that both SRA and CSTA are critical for performance. The introduction of SRA alone provides the most significant leap, dramatically improving IoU from 0.503 to 0.554 and slashing the MMD from 0.116 to 0.065. The subsequent addition of CSTA further enhances performance across all metrics, leading to the best overall scores, including a final IoU of 0.583. This clearly shows that while SRA captures the core geometric structure, CSTA is essential for achieving the highest level of temporal and distributional fidelity.

\begin{table}
\centering
\caption{\textbf{Combining attention mechanisms in START yields superior performance.} The table ablates our two attention mechanisms, showing that each provides a substantial gain over the baseline. Their combination in the START module is most effective, for instance, improving IoU from 0.503 to 0.583. Bold denotes the best performance.}
\label{tab:ab3}
\scalebox{0.9}{
\begin{tabular}{cc|cccc}
\toprule
\textbf{SRA} & \textbf{CSTA} & {IoU$\uparrow$} & {Chamfer$\downarrow$} & {MMD ($10^{-4}$)$\downarrow$} & {JSD$\downarrow$} \\
\midrule
& & 0.503 & 0.021 & 0.116 & 0.061 \\
\checkmark & & 0.554 & 0.020 & 0.065 &  0.060\\
\checkmark & \checkmark & \textbf{0.583} & \textbf{0.017}& \textbf{0.061} & \textbf{0.056} \\
\bottomrule
\end{tabular}
}
\end{table}



\subsection{Qualitative Results}


Our qualitative results in Fig. \ref{fig:rec_result} and \ref{fig:pre_gen} demonstrate LiSTAR's clear superiority across all tasks. In reconstruction, the baseline accumulates significant artifacts (blue points) over time, while our method maintains high fidelity with more true positives (magenta). Similarly, for prediction and generation, the baseline's outputs become progressively blurry and lose structural detail, whereas our results remain sharp and temporally consistent, closely matching the ground truth. This visual evidence confirms our model's advanced capability in modeling complex 4D dynamics with high fidelity. Further experimental details are provided in the Appendix.



\section{Conclusion}



In this paper, we introduced LiSTAR, a novel generative world model for high-fidelity, controllable 4D LiDAR synthesis. By unifying a novel HCS representation with a START, LiSTAR effectively preserves geometric fidelity and ensures temporal coherence. Its discrete MaskSTART framework further enables efficient, high-resolution generation conditioned on scene layouts. We have demonstrated that LiSTAR establishes a new state-of-the-art across reconstruction, prediction, and generation tasks, providing a powerful tool for creating realistic simulation environments for autonomous driving. Future work could explore multi-modal conditioning for even richer scene synthesis.

\clearpage
{
    \small
    \bibliographystyle{ieeenat_fullname}
    \bibliography{main}

@String(AAAI = {AAAI})

@article{hu2023gaia,
  title={{GAIA-1: A Generative World Model for Autonomous Driving}},
  author={Hu, Anthony and Russell, Lloyd and Yeo, Hudson and Murez, Zak and Fedoseev, George and Kendall, Alex and Shotton, Jamie and Corrado, Gianluca},
  journal={arXiv preprint arXiv:2309.17080},
  year={2023}
}

@article{mei2024dreamforge,
  title={{DreamForge: Motion-Aware Autoregressive Video Generation for Multi-View Driving Scenes}},
  author={Mei, Jianbiao and Hu, Tao and Yang, Xuemeng and Wen, Licheng and Yang, Yu and Wei, Tiantian and Ma, Yukai and Dou, Min and Shi, Botian and Liu, Yong},
  journal={arXiv preprint arXiv:2409.04003},
  year={2024}
}

@article{gao2023magicdrive,
  title={{MagicDrive: Street View Generation with Diverse 3D Geometry Control}},
  author={Gao, Ruiyuan and Chen, Kai and Xie, Enze and Hong, Lanqing and Li, Zhenguo and Yeung, Dit-Yan and Xu, Qiang},
  journal={arXiv preprint arXiv:2310.02601},
  year={2023}
}

@article{wang2024occsora,
  title={{OccSora: 4D Occupancy Generation Models as World Simulators for Autonomous Driving}},
  author={Wang, Lening and Zheng, Wenzhao and Ren, Yilong and Jiang, Han and Cui, Zhiyong and Yu, Haiyang and Lu, Jiwen},
  journal={arXiv preprint arXiv:2405.20337},
  year={2024}
}

@inproceedings{kong2023robo3d,
  title={{Robo3D: Towards Robust and Reliable 3D Perception against Corruptions}},
  author={Kong, Lingdong and Liu, Youquan and Li, Xin and Chen, Runnan and Zhang, Wenwei and Ren, Jiawei and Pan, Liang and Chen, Kai and Liu, Ziwei},
  booktitle={Proceedings of the IEEE/CVF International Conference on Computer Vision},
  pages={19994--20006},
  year={2023}
}

@article{liu2023segment,
  title={{Segment Any Point Cloud Sequences by Distilling Vision Foundation Models}},
  author={Liu, Youquan and Kong, Lingdong and Cen, Jun and Chen, Runnan and Zhang, Wenwei and Pan, Liang and Chen, Kai and Liu, Ziwei},
  journal={Advances in Neural Information Processing Systems},
  volume={36},
  pages={37193--37229},
  year={2023}
}

@inproceedings{xu20244d,
  title={{4D Contrastive Superflows are Dense 3D Representation Learners}},
  author={Xu, Xiang and Kong, Lingdong and Shuai, Hui and Zhang, Wenwei and Pan, Liang and Chen, Kai and Liu, Ziwei and Liu, Qingshan},
  booktitle={European Conference on Computer Vision},
  pages={58--80},
  year={2024},
  organization={Springer}
}

@inproceedings{zyrianov2022learning,
  title={{Learning to Generate Realistic LiDAR Point Clouds}},
  author={Zyrianov, Vlas and Zhu, Xiyue and Wang, Shenlong},
  booktitle={European Conference on Computer Vision},
  pages={17--35},
  year={2022}
}

@article{li2023pointdmm,
  title={{PointDMM: A Deep-Learning-Based Semantic Segmentation Method for Point Clouds in Complex Forest Environments}},
  author={Li, Jiang and Liu, Jinhao and Huang, Qingqing},
  journal={Forests},
  volume={14},
  number={12},
  pages={2276},
  year={2023},
  publisher={MDPI}
}

@inproceedings{yu2023video,
  title={{Video Probabilistic Diffusion Models in Projected Latent Space}},
  author={Yu, Sihyun and Sohn, Kihyuk and Kim, Subin and Shin, Jinwoo},
  booktitle={Proceedings of the IEEE/CVF conference on computer vision and pattern recognition},
  pages={18456--18466},
  year={2023}
}

@article{zhang2023copilot4d,
  title={{Copilot4D: Learning Unsupervised World Models for Autonomous Driving via Discrete Diffusion}},
  author={Zhang, Lunjun and Xiong, Yuwen and Yang, Ze and Casas, Sergio and Hu, Rui and Urtasun, Raquel},
  journal={arXiv preprint arXiv:2311.01017},
  year={2023}
}

@article{liang2025lidarcrafter,
  title={{LiDARCrafter: Dynamic 4D World Modeling from LiDAR Sequences}},
  author={Liang, Ao and Liu, Youquan and Yang, Yu and Lu, Dongyue and Li, Linfeng and Kong, Lingdong and Zhao, Huaici and Ooi, Wei Tsang},
  journal={arXiv preprint arXiv:2508.03692},
  year={2025}
}

@inproceedings{li2025uniscene,
  title={{UniScene: Unified Occupancy-centric Driving Scene Generation}},
  author={Li, Bohan and Guo, Jiazhe and Liu, Hongsi and Zou, Yingshuang and Ding, Yikang and Chen, Xiwu and Zhu, Hu and Tan, Feiyang and Zhang, Chi and Wang, Tiancai and others},
  booktitle={Proceedings of the Computer Vision and Pattern Recognition Conference},
  pages={11971--11981},
  year={2025}
}

@inproceedings{huang2024vbench,
  title={{VBench: Comprehensive Benchmark Suite for Video Generative Models}},
  author={Huang, Ziqi and He, Yinan and Yu, Jiashuo and Zhang, Fan and Si, Chenyang and Jiang, Yuming and Zhang, Yuanhan and Wu, Tianxing and Jin, Qingyang and Chanpaisit, Nattapol and others},
  booktitle={Proceedings of the IEEE/CVF Conference on Computer Vision and Pattern Recognition},
  pages={21807--21818},
  year={2024}
}

@inproceedings{nakashima2024lidar,
  title={{LiDAR Data Synthesis with Denoising Diffusion Probabilistic Models}},
  author={Nakashima, Kazuto and Kurazume, Ryo},
  booktitle={2024 IEEE International Conference on Robotics and Automation (ICRA)},
  pages={14724--14731},
  year={2024}
}

@inproceedings{nakashima2025fast,
  title={{Fast LiDAR Data Generation with Rectified Flows}},
  author={Nakashima, Kazuto and Liu, Xiaowen and Miyawaki, Tomoya and Iwashita, Yumi and Kurazume, Ryo},
  booktitle={2025 IEEE International Conference on Robotics and Automation (ICRA)},
  pages={10057--10063},
  year={2025},
  organization={IEEE}
}

@article{lin2017structured,
  title={{A Structured Self-attentive Sentence Embedding}},
  author={Lin, Zhouhan and Feng, Minwei and Santos, Cicero Nogueira dos and Yu, Mo and Xiang, Bing and Zhou, Bowen and Bengio, Yoshua},
  journal={arXiv preprint arXiv:1703.03130},
  year={2017}
}

@inproceedings{qi2017pointnet,
  title={{PointNet: Deep Learning on Point Sets for 3D Classification and Segmentation}},
  author={Qi, Charles R and Su, Hao and Mo, Kaichun and Guibas, Leonidas J},
  booktitle={Proceedings of the IEEE conference on computer vision and pattern recognition},
  pages={652--660},
  year={2017}
}

@article{qi2017pointnet++,
  title={{PointNet++: Deep Hierarchical Feature Learning on Point Sets in a Metric Space}},
  author={Qi, Charles Ruizhongtai and Yi, Li and Su, Hao and Guibas, Leonidas J},
  journal={Advances in neural information processing systems},
  volume={30},
  year={2017}
}

@inproceedings{qi2016volumetric,
  title={{Volumetric and Multi-View CNNs for Object Classification on 3D Data}},
  author={Qi, Charles R and Su, Hao and Nie{\ss}ner, Matthias and Dai, Angela and Yan, Mengyuan and Guibas, Leonidas J},
  booktitle={Proceedings of the IEEE conference on computer vision and pattern recognition},
  pages={5648--5656},
  year={2016}
}

@inproceedings{yang2018pixor,
  title={{PIXOR: Real-Time 3D Object Detection From Point Clouds}},
  author={Yang, Bin and Luo, Wenjie and Urtasun, Raquel},
  booktitle={Proceedings of the IEEE conference on Computer Vision and Pattern Recognition},
  pages={7652--7660},
  year={2018}
}

@article{liu2021pvnas,
  title={{PVNAS: 3D Neural Architecture Search With Point-Voxel Convolution}},
  author={Liu, Zhijian and Tang, Haotian and Zhao, Shengyu and Shao, Kevin and Han, Song},
  journal={IEEE Transactions on Pattern Analysis and Machine Intelligence},
  volume={44},
  number={11},
  pages={8552--8568},
  year={2021}
}

@inproceedings{choy20163d,
  title={{3D-R2N2: A Unified Approach for Single and Multi-view 3D Object Reconstruction}},
  author={Choy, Christopher B and Xu, Danfei and Gwak, JunYoung and Chen, Kevin and Savarese, Silvio},
  booktitle={European conference on computer vision},
  pages={628--644},
  year={2016}
}

@inproceedings{graham20183d,
  title={{3D Semantic Segmentation With Submanifold Sparse Convolutional Networks}},
  author={Graham, Benjamin and Engelcke, Martin and Van Der Maaten, Laurens},
  booktitle={Proceedings of the IEEE conference on computer vision and pattern recognition},
  pages={9224--9232},
  year={2018}
}

@inproceedings{lang2019pointpillars,
  title={{PointPillars: Fast Encoders for Object Detection From Point Clouds}},
  author={Lang, Alex H and Vora, Sourabh and Caesar, Holger and Zhou, Lubing and Yang, Jiong and Beijbom, Oscar},
  booktitle={Proceedings of the IEEE/CVF conference on computer vision and pattern recognition},
  pages={12697--12705},
  year={2019}
}

@inproceedings{hu2024rangeldm,
  title={{RangeLDM: Fast Realistic LiDAR Point Cloud Generation}},
  author={Hu, Qianjiang and Zhang, Zhimin and Hu, Wei},
  booktitle={European Conference on Computer Vision},
  pages={115--135},
  year={2024}
}

@inproceedings{milioto2019rangenet++,
  title={{RangeNet ++: Fast and Accurate LiDAR Semantic Segmentation}},
  author={Milioto, Andres and Vizzo, Ignacio and Behley, Jens and Stachniss, Cyrill},
  booktitle={2019 IEEE/RSJ international conference on intelligent robots and systems (IROS)},
  pages={4213--4220},
  year={2019}
}

@inproceedings{ran2024towards,
  title={{Towards Realistic Scene Generation with LiDAR Diffusion Models}},
  author={Ran, Haoxi and Guizilini, Vitor and Wang, Yue},
  booktitle={Proceedings of the IEEE/CVF Conference on Computer Vision and Pattern Recognition},
  pages={14738--14748},
  year={2024}
}

@article{kingma2013auto,
  title={{Auto-Encoding Variational Bayes}},
  author={Kingma, Diederik P and Welling, Max},
  journal={arXiv preprint arXiv:1312.6114},
  year={2013}
}

@article{van2017neural,
  title={{Neural Discrete Representation Learning}},
  author={Van Den Oord, Aaron and Vinyals, Oriol and others},
  journal={Advances in neural information processing systems},
  volume={30},
  year={2017}
}

@inproceedings{esser2021taming,
  title={{Taming Transformers for High-Resolution Image Synthesis}},
  author={Esser, Patrick and Rombach, Robin and Ommer, Bjorn},
  booktitle={Proceedings of the IEEE/CVF conference on computer vision and pattern recognition},
  pages={12873--12883},
  year={2021}
}

@inproceedings{caccia2019deep,
  title={{Deep Generative Modeling of LiDAR Data}},
  author={Caccia, Lucas and Van Hoof, Herke and Courville, Aaron and Pineau, Joelle},
  booktitle={2019 IEEE/RSJ International Conference on Intelligent Robots and Systems (IROS)},
  pages={5034--5040},
  year={2019}
}

@inproceedings{amini2022vista,
  title={{VISTA 2.0: An Open, Data-driven Simulator for Multimodal Sensing and Policy Learning for Autonomous Vehicles}},
  author={Amini, Alexander and Wang, Tsun-Hsuan and Gilitschenski, Igor and Schwarting, Wilko and Liu, Zhijian and Han, Song and Karaman, Sertac and Rus, Daniela},
  booktitle={2022 International Conference on Robotics and Automation (ICRA)},
  pages={2419--2426},
  year={2022}
}

@inproceedings{dosovitskiy2017carla,
  title={{CARLA: An Open Urban Driving Simulator}},
  author={Dosovitskiy, Alexey and Ros, German and Codevilla, Felipe and Lopez, Antonio and Koltun, Vladlen},
  booktitle={Conference on robot learning},
  pages={1--16},
  year={2017}
}

@inproceedings{manivasagam2020lidarsim,
  title={{LiDARsim: Realistic LiDAR Simulation by Leveraging the Real World}},
  author={Manivasagam, Sivabalan and Wang, Shenlong and Wong, Kelvin and Zeng, Wenyuan and Sazanovich, Mikita and Tan, Shuhan and Yang, Bin and Ma, Wei-Chiu and Urtasun, Raquel},
  booktitle={Proceedings of the IEEE/CVF Conference on Computer Vision and Pattern Recognition},
  pages={11167--11176},
  year={2020}
}

@inproceedings{nakashima2021learning,
  title={{Learning to Drop Points for LiDAR Scan Synthesis}},
  author={Nakashima, Kazuto and Kurazume, Ryo},
  booktitle={2021 IEEE/RSJ International Conference on Intelligent Robots and Systems (IROS)},
  pages={222--229},
  year={2021}
}

@article{mildenhall2021nerf,
  title={{NeRF: Representing Scenes
as Neural Radiance Fields
for View Synthesis}},
  author={Mildenhall, Ben and Srinivasan, Pratul P and Tancik, Matthew and Barron, Jonathan T and Ramamoorthi, Ravi and Ng, Ren},
  journal={Communications of the ACM},
  volume={65},
  number={1},
  pages={99--106},
  year={2021}
}

@inproceedings{zhang2024nerf,
  title={{NeRF-LiDAR: Generating Realistic LiDAR Point Clouds with Neural Radiance Fields}},
  author={Zhang, Junge and Zhang, Feihu and Kuang, Shaochen and Zhang, Li},
  booktitle={Proceedings of the AAAI Conference on Artificial Intelligence},
  volume={38},
  number={7},
  pages={7178--7186},
  year={2024}
}

@article{russell2025gaia,
  title={{GAIA-2: A Controllable Multi-View Generative World Model for Autonomous Driving}},
  author={Russell, Lloyd and Hu, Anthony and Bertoni, Lorenzo and Fedoseev, George and Shotton, Jamie and Arani, Elahe and Corrado, Gianluca},
  journal={arXiv preprint arXiv:2503.20523},
  year={2025}
}

@inproceedings{zhao2025drivedreamer,
  title={{DriveDreamer-2: LLM-Enhanced World Models for Diverse Driving Video Generation}},
  author={Zhao, Guosheng and Wang, Xiaofeng and Zhu, Zheng and Chen, Xinze and Huang, Guan and Bao, Xiaoyi and Wang, Xingang},
  booktitle={Proceedings of the AAAI Conference on Artificial Intelligence},
  volume={39},
  number={10},
  pages={10412--10420},
  year={2025}
}

@inproceedings{zhao2025drivedreamer4d,
  title={{DriveDreamer4D: World Models Are Effective Data Machines for 4D Driving Scene Representation}},
  author={Zhao, Guosheng and Ni, Chaojun and Wang, Xiaofeng and Zhu, Zheng and Zhang, Xueyang and Wang, Yida and Huang, Guan and Chen, Xinze and Wang, Boyuan and Zhang, Youyi and others},
  booktitle={Proceedings of the Computer Vision and Pattern Recognition Conference},
  pages={12015--12026},
  year={2025}
}

@article{wu2024drivescape,
  title={{DriveScape: Towards High-Resolution Controllable Multi-View Driving Video Generation}},
  author={Wu, Wei and Guo, Xi and Tang, Weixuan and Huang, Tingxuan and Wang, Chiyu and Chen, Dongyue and Ding, Chenjing},
  journal={arXiv preprint arXiv:2409.05463},
  year={2024}
}

@inproceedings{zheng2024occworld,
  title={{OccWorld: Learning a 3D Occupancy World Model for Autonomous Driving}},
  author={Zheng, Wenzhao and Chen, Weiliang and Huang, Yuanhui and Zhang, Borui and Duan, Yueqi and Lu, Jiwen},
  booktitle={European conference on computer vision},
  pages={55--72},
  year={2024}
}

@article{wei2024occllama,
  title={{OccLLaMA: An Occupancy-Language-Action Generative World Model for Autonomous Driving}},
  author={Wei, Julong and Yuan, Shanshuai and Li, Pengfei and Hu, Qingda and Gan, Zhongxue and Ding, Wenchao},
  journal={arXiv preprint arXiv:2409.03272},
  year={2024}
}

@inproceedings{zuo2025gaussianworld,
  title={{GaussianWorld: Gaussian World Model for Streaming 3D Occupancy Prediction}},
  author={Zuo, Sicheng and Zheng, Wenzhao and Huang, Yuanhui and Zhou, Jie and Lu, Jiwen},
  booktitle={Proceedings of the Computer Vision and Pattern Recognition Conference},
  pages={6772--6781},
  year={2025}
}

@inproceedings{li2025occmamba,
  title={{OccMamba: Semantic Occupancy Prediction with State Space Models}},
  author={Li, Heng and Hou, Yuenan and Xing, Xiaohan and Ma, Yuexin and Sun, Xiao and Zhang, Yanyong},
  booktitle={Proceedings of the Computer Vision and Pattern Recognition Conference},
  pages={11949--11959},
  year={2025}
}

@article{gu2024dome,
  title={{DOME: Taming Diffusion Model into High-Fidelity Controllable Occupancy World Model}},
  author={Gu, Songen and Yin, Wei and Jin, Bu and Guo, Xiaoyang and Wang, Junming and Li, Haodong and Zhang, Qian and Long, Xiaoxiao},
  journal={arXiv preprint arXiv:2410.10429},
  year={2024}
}

@inproceedings{zyrianov2025lidardm,
  title={{LidarDM: Generative LiDAR Simulation in a Generated World}},
  author={Zyrianov, Vlas and Che, Henry and Liu, Zhijian and Wang, Shenlong},
  booktitle={2025 IEEE International Conference on Robotics and Automation (ICRA)},
  pages={6055--6062},
  year={2025}
}

@inproceedings{yang2024visual,
  title={{Visual Point Cloud Forecasting enables Scalable Autonomous Driving}},
  author={Yang, Zetong and Chen, Li and Sun, Yanan and Li, Hongyang},
  booktitle={Proceedings of the IEEE/CVF Conference on Computer Vision and Pattern Recognition},
  pages={14673--14684},
  year={2024}
}

@article{ho2020denoising,
  title={{Denoising Diffusion Probabilistic Models}},
  author={Ho, Jonathan and Jain, Ajay and Abbeel, Pieter},
  journal={Advances in neural information processing systems},
  volume={33},
  pages={6840--6851},
  year={2020}
}

@article{song2020denoising,
  title={{Denoising Diffusion Implicit Models}},
  author={Song, Jiaming and Meng, Chenlin and Ermon, Stefano},
  journal={arXiv preprint arXiv:2010.02502},
  year={2020}
}

@article{song2020score,
  title={{Score-Based Generative Modeling through Stochastic Differential Equations}},
  author={Song, Yang and Sohl-Dickstein, Jascha and Kingma, Diederik P and Kumar, Abhishek and Ermon, Stefano and Poole, Ben},
  journal={arXiv preprint arXiv:2011.13456},
  year={2020}
}

@inproceedings{rombach2022high,
  title={{High-Resolution Image Synthesis With Latent Diffusion Models}},
  author={Rombach, Robin and Blattmann, Andreas and Lorenz, Dominik and Esser, Patrick and Ommer, Bj{\"o}rn},
  booktitle={Proceedings of the IEEE/CVF conference on computer vision and pattern recognition},
  pages={10684--10695},
  year={2022}
}

@inproceedings{chang2022maskgit,
  title={{MaskGIT: Masked Generative Image Transformer}},
  author={Chang, Huiwen and Zhang, Han and Jiang, Lu and Liu, Ce and Freeman, William T},
  booktitle={Proceedings of the IEEE/CVF conference on computer vision and pattern recognition},
  pages={11315--11325},
  year={2022}
}

@inproceedings{devlin2019bert,
  title={{BERT: Pre-training of Deep Bidirectional Transformers for Language Understanding}},
  author={Devlin, Jacob and Chang, Ming-Wei and Lee, Kenton and Toutanova, Kristina},
  booktitle={Proceedings of the 2019 conference of the North American chapter of the association for computational linguistics: human language technologies, volume 1 (long and short papers)},
  pages={4171--4186},
  year={2019}
}

@inproceedings{yu2023magvit,
  title={{MAGVIT: Masked Generative Video Transformer}},
  author={Yu, Lijun and Cheng, Yong and Sohn, Kihyuk and Lezama, Jos{\'e} and Zhang, Han and Chang, Huiwen and Hauptmann, Alexander G and Yang, Ming-Hsuan and Hao, Yuan and Essa, Irfan and others},
  booktitle={Proceedings of the IEEE/CVF Conference on Computer Vision and Pattern Recognition},
  pages={10459--10469},
  year={2023}
}

@inproceedings{xiong2023learning,
  title={{Learning Compact Representations for LiDAR Completion and Generation}},
  author={Xiong, Yuwen and Ma, Wei-Chiu and Wang, Jingkang and Urtasun, Raquel},
  booktitle={Proceedings of the IEEE/CVF Conference on Computer Vision and Pattern Recognition},
  pages={1074--1083},
  year={2023}
}

@article{lipman2022flow,
  title={{Flow Matching for Generative Modeling}},
  author={Lipman, Yaron and Chen, Ricky TQ and Ben-Hamu, Heli and Nickel, Maximilian and Le, Matt},
  journal={arXiv preprint arXiv:2210.02747},
  year={2022}
}

@article{geng2025mean,
  title={{Mean Flows for One-step Generative Modeling}},
  author={Geng, Zhengyang and Deng, Mingyang and Bai, Xingjian and Kolter, J Zico and He, Kaiming},
  journal={arXiv preprint arXiv:2505.13447},
  year={2025}
}

@inproceedings{liu2021swin,
  title={{Swin Transformer: Hierarchical Vision Transformer Using Shifted Windows}},
  author={Liu, Ze and Lin, Yutong and Cao, Yue and Hu, Han and Wei, Yixuan and Zhang, Zheng and Lin, Stephen and Guo, Baining},
  booktitle={Proceedings of the IEEE/CVF international conference on computer vision},
  pages={10012--10022},
  year={2021}
}

@misc{opendwm,
  Year = {2025},
  Note = {https://github.com/SenseTime-FVG/OpenDWM},
  Title = {OpenDWM: Open Driving World Models}
}

@article{chen2024unimlvg,
  title={UniMLVG: Unified Framework for Multi-view Long Video Generation with Comprehensive Control Capabilities for Autonomous Driving},
  author={Chen, Rui and Wu, Zehuan and Liu, Yichen and Guo, Yuxin and Ni, Jingcheng and Xia, Haifeng and Xia, Siyu},
  journal={arXiv preprint arXiv:2412.04842},
  year={2024}
}

@article{ni2025maskgwm,
  title={MaskGWM: A Generalizable Driving World Model with Video Mask Reconstruction},
  author={Ni, Jingcheng and Guo, Yuxin and Liu, Yichen and Chen, Rui and Lu, Lewei and Wu, Zehuan},
  journal={arXiv preprint arXiv:2502.11663},
  year={2025}
}

@inproceedings{weng2021inverting,
  title={{Inverting the Pose Forecasting Pipeline with SPF2: Sequential Pointcloud Forecasting for Sequential Pose Forecasting}},
  author={Weng, Xinshuo and Wang, Jianren and Levine, Sergey and Kitani, Kris and Rhinehart, Nicholas},
  booktitle={Conference on robot learning},
  pages={11--20},
  year={2021},
  organization={PMLR}
}

@inproceedings{weng2022s2net,
  title={{S2Net: Stochastic Sequential Pointcloud Forecasting}},
  author={Weng, Xinshuo and Nan, Junyu and Lee, Kuan-Hui and McAllister, Rowan and Gaidon, Adrien and Rhinehart, Nicholas and Kitani, Kris M},
  booktitle={European Conference on Computer Vision},
  pages={549--564},
  year={2022},
  organization={Springer}
}

@inproceedings{khurana2023point,
  title={{Point Cloud Forecasting as a Proxy for 4D Occupancy Forecasting}},
  author={Khurana, Tarasha and Hu, Peiyun and Held, David and Ramanan, Deva},
  booktitle={Proceedings of the IEEE/CVF Conference on Computer Vision and Pattern Recognition},
  pages={1116--1124},
  year={2023}
}

@inproceedings{caesar2020nuscenes,
  title={{nuScenes: A Multimodal Dataset for Autonomous Driving}},
  author={Caesar, Holger and Bankiti, Varun and Lang, Alex H and Vora, Sourabh and Liong, Venice Erin and Xu, Qiang and Krishnan, Anush and Pan, Yu and Baldan, Giancarlo and Beijbom, Oscar},
  booktitle={Proceedings of the IEEE/CVF conference on computer vision and pattern recognition},
  pages={11621--11631},
  year={2020}
}
}

\clearpage
\setcounter{page}{1}
\maketitlesupplementary

\section{Methodology}
\label{sec:append1}

\subsection{HCS-based 4D VQ-VAE}

To capture both geometric structures and temporal dynamics inherent in sequential LiDAR scans, we design a VQ-VAE tailored to the 4D HCS voxel representation. This framework effectively abstracts redundant measurements while preserving crucial spatiotemporal information, enabling reconstruction, future sequence prediction and generation. The framework consists of a hierarchical encoder that maps the input sequence to a discrete latent space and a generative decoder that reconstructs the 4D volume from this representation.

\subsubsection{Hierarchical Encoder}
The encoder $E$, detailed in Alg.~\ref{alg:encoder}, transforms the input sequence $\mathbf{x} \in \mathbb{R}^{\rho \times \theta \times \phi \times T}$ into a compact latent representation $\mathbf{z} = E(\mathbf{x}) \in \mathbb{R}^{x \times x \times x \times D}$. This process begins with spherical coordinate voxelization to obtain voxel features $\{v_1, v_2, \dots, v_N\}$, which addresses the non-uniform distribution problem inherent to point clouds across varying viewing angles and distances. The voxel feature encoding employs our CSTA module for comprehensive cross-dimensional interaction, allowing features to be processed across both spatial and temporal domains. Subsequently, the SRA module enhances feature correlations among voxels along shared laser-ray directions, fostering coherent representation.

Inspired by video generation advances, our 4D-VAE model enables unified spatiotemporal processing, avoiding frame-by-frame limitations that often compromise temporal consistency. Encoding involves patch merging of voxels, followed by the application of four stacked START blocks to extract spatiotemporal features. A 
$2\times2\times1$ downsampling operation is applied, reducing feature representation by a factor of $8\times8\times2$ in polar BEV space. Further encoding proceeds through START blocks without ray-specific attention, ultimately generating a robust latent space 
$z \in \mathbb{R}^{X \times X \times X \times D}$, which undergoes vector quantization yielding $\hat{z} \in \mathcal{C}$, where $\mathcal{C}$ is the codebook of latent vectors.

\begin{algorithm}
\caption{Encoder of HCS-based 4D VQ-VAE}
\label{alg:encoder}
\begin{algorithmic}[1]
\Input Input voxelized LiDAR sequence $V \in \mathbb{R}^{\rho \times \theta \times \phi \times T}$.
\Output Latent representation $z \in \mathbb{R}^{H' \times W' \times D' \times C'}$.

\State $h \gets \text{PatchMerge}(V)$ \Comment{Initial patch merging of voxels}
\For{$i=1 \to 4$} \Comment{Apply four stacked START blocks for spatiotemporal feature extraction}
    \State $h \gets \text{START\_Block}_i(h)$
\EndFor
\State $h \gets \text{Downsample}_{2\times2\times1}(h)$ \Comment{Reduce spatial resolution}
\For{$i=1 \to K$} \Comment{Further encoding with CSTA blocks}
    \State $h \gets \text{CSTA\_Block}_i(h)$
\EndFor
\State $z \gets h$ \Comment{Final latent representation}
\State \Return $z$
\end{algorithmic}
\end{algorithm}

\subsubsection{Generative Decoder}
The decoder $D$ reconstructs the LiDAR volume from the quantized latent representation $\hat{z} \in \mathcal{C}$, as outlined in Alg.~\ref{alg:decoder}. producing $\tilde{\mathbf{x}} = D(\hat{z}) \in \mathbb{R}^{\rho \times \theta \times \phi \times T}$, ensuring both geometric fidelity and temporal coherence. Initial processing leverages eight STA blocks without Ray Attention for feature restoration, followed by a 
$2\times2\times1$ upsampling operation. Next, two START blocks refine spatiotemporal information and depth cues along ray directions, reinforcing structural integrity and continuity within the reconstructed output. The features are then upsampled back to the original voxel size, with the final point cloud $\tilde{\mathcal{P}} = \{ \tilde{p}_1, \tilde{p}_2, \dots, \tilde{p}_M \}$ rendered by a dedicated module, ensuring consistent geometry and smooth motion trajectories.

\begin{algorithm}
\caption{Decoder of HCS-based 4D VQ-VAE}
\label{alg:decoder}
\begin{algorithmic}[1]
\Input Quantized latent representation $\hat{z} \in \mathbb{R}^{H' \times W' \times D' \times C'}$.
\Output Reconstructed voxel volume $V_{\text{out}}$ and point cloud $\tilde{\mathcal{P}}$.

\State $h' \gets \hat{z}$ \Comment{Start decoding from quantized latent}
\For{$i=1 \to 8$} \Comment{Initial feature restoration with STA blocks}
    \State $h' \gets \text{STA\_Block}_i(h')$
\EndFor
\State $h' \gets \text{Upsample}_{2\times2\times1}(h')$ \Comment{Increase spatial resolution}
\For{$i=1 \to 2$} \Comment{Refine with START blocks to reinforce ray structure}
    \State $h' \gets \text{START\_Block}_i(h')$
\EndFor
\State $V_{\text{out}} \gets \text{Upsample}(h')$ \Comment{Upsample to original voxel resolution}
\State $\tilde{\mathcal{P}} \gets \text{RenderModule}(V_{\text{out}})$ \Comment{Render final point cloud from voxels}
\State \Return $V_{\text{out}}, \tilde{\mathcal{P}}$
\end{algorithmic}
\end{algorithm}

\subsubsection{Loss Function}
The encoder-decoder framework is trained by minimizing a loss function with three components: vector quantization loss, voxel reconstruction loss, and point cloud reconstruction loss:
\begin{equation}
    L=L_{VQ}(z,\hat{z})+L_{v}(V_{pred},V_{target})+L_{p}(U_{pred},V_{target}),
\end{equation}
where $z$ and $\hat{z}$ are encoder outputs and quantized features, respectively; $V_{pred}$ denotes the predicted voxels, $U_{pred}$ represents voxelized rendered point cloud; and $V_{target}$ denotes target voxels. This loss formulation ensures fidelity across both voxel representation and rendered point clouds, enhancing reconstruction quality.

\subsection{MaskSTART Module}
\begin{figure}[!ht] 
    \centering
    \includegraphics[width=0.5\textwidth]{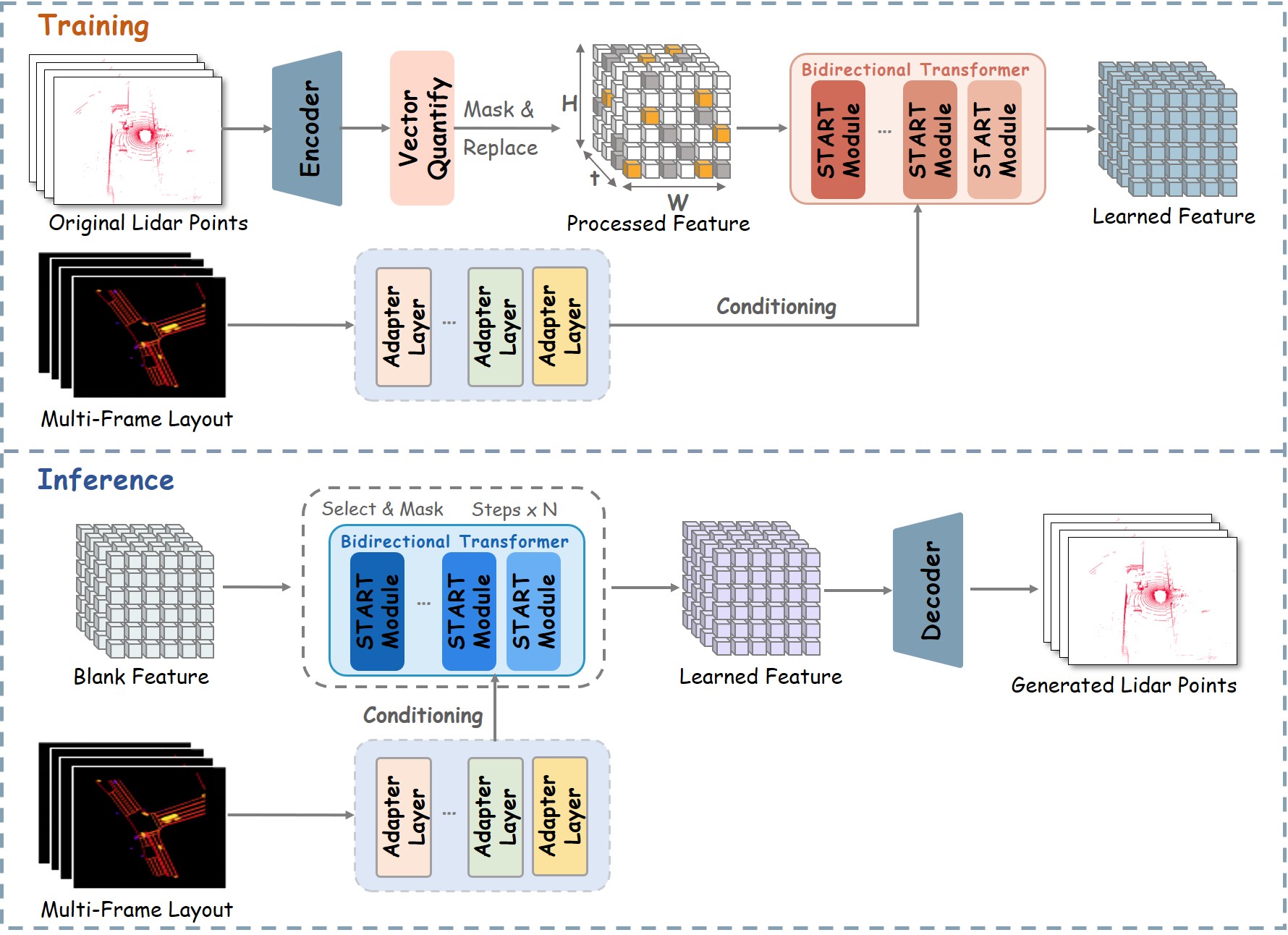}
    \caption{Generation training and inference} 
    \label{fig:figure_train}
\end{figure}

We present the MaskSTART Module, a comprehensive framework designed for both point cloud prediction and generation tasks. Illustrated in Fig.~\ref{fig:figure_train}, this module excels in generating future LiDAR sequences $\{\tilde{\mathcal{P}}_{t+1}, \dots, \tilde{\mathcal{P}}_{t+\tau}\}$ from past observations $\{\mathcal{P}_{t-\tau+1}, \dots, \mathcal{P}_{t}\}$ while incorporating various conditional inputs for generation, such as scene layouts, textual descriptions, or visual cues.

Initially, the method aligns raw point cloud coordinates $\mathcal{P}_t = \{p_i\}_{i=1}^M $ with the LiDAR's inherent geometric distribution through conversion to spherical coordinates$(\rho, \theta, \phi)$, followed by voxelization into a grid $\mathbf{x}_t \in \mathbb{R}^{\rho \times \theta \times \phi}$. A tokenizer $T(\cdot)$ then maps the voxel grid into a discrete token sequence $\mathbf{s}_t = T(\mathbf{x}_t)$, which is processed by a MaskSTART module for masked generative tasks under differing conditional settings.

\subsubsection{Regional Spatiotemporal Attention}

The MaskSTART module integrates $N$ stacked Transformer blocks designed to address spatiotemporal inconsistencies arising from the motion of the ego-vehicle and dynamic objects. These inconsistencies can compromise the temporal coherence of generated point clouds in long sequences.

To effectively capture long-range dependencies and resolve these issues, we introduce Regional Spatio-Temporal Attention (RSTA). Unlike conventional attention mechanisms that compute attention scores across all tokens $\mathbf{s}_t = \{s_{t,1}, \dots, s_{t,M}\}$, RSTA predicts offsets $\Delta_{t,i}$ for each token $s_{t,i}$ to locate regions of interest (RoIs) $\mathcal{R}_{t,i} \subseteq \mathbf{s}_{\leq t}$, facilitating attention computation exclusively within these targeted areas. This selective approach efficiently models spatiotemporal variations while minimizing computational overhead.

RSTA maintains strict causality by preserving the chronological progression of events, ensuring predictions rely solely on past and present information. This causal framework, paired with the 3D Swin Transformer's Temporal Attention, provides enhanced temporal consistency and spatial fidelity in point cloud sequences, making RSTA particularly suited for handling dynamic environments.

\subsubsection{Prediction Task}
In prediction tasks, the model takes as input a sequence of historical observations ${\{h^1,...,h^t\}}$, where each observation ${h^t = \{o^t, c^t\}}$ consists of LiDAR point cloud data $o_t$  and associated conditional information $c_t$ (e.g., ego-vehicle pose). The goal of the model is to learn a probabilistic world model $p_\theta$ capable of predicting future point clouds conditioned on the historical context. During training, as detailed in Alg.~\ref{alg:training_prediction},the model encodes the input point cloud sequence using an hierarchical  encoder to extract latent features. Historical frames are encoded and serve as the conditional input, while future frame tokens are processed with a masking strategy to improve robustness and generation accuracy. Specifically, tokens in future frames are either randomly replaced with other tokens from the codebook, assigned learnable mask tokens, or left unaltered. The MaskSTART module learns to infer the masked tokens based on the given historical sequence, optimizing the following loss function:

\begin{equation}
\label{eq:masked_prediction_loss}
L_{\text{CE}} = - \sum_{i \in \mathcal{M}} \log p(z_{q,i} | z'_{q})
\end{equation}
$L_{\text{CE}}$ is the total Cross-Entropy loss.
$M$ is the set of indices corresponding to all masked token positions in the future sequence. The loss is computed exclusively over these positions.
$z_{q,i}$ is the ground-truth token (i.e., the correct codebook index) at position $i$. $z'_{q}$ is the masked input sequence provided to the MaskSTART module, which consists of the historical context and the corrupted future sequence. $p(z_{q,i} | z'_{q})$ is the predicted probability.

\begin{algorithm}[t]
\caption{Training for Prediction with MaskSTART}
\label{alg:training_prediction}
\begin{algorithmic}[1]
\Input Historical observation sequence $O_{\text{hist}} = \{o^1, ..., o^t\}$.
\Input Future observation sequence $O_{\text{future}} = \{o^{t+1}, ..., o^{T}\}$.
\Input Encoder $E$, Codebook $\mathcal{C}$, MaskSTART module $M_{\text{START}}$.
\Output Trained parameters for $M_{\text{START}}$.

\State \Comment{1. Encode full sequence and get ground-truth tokens}
\State $O_{\text{full}} \gets \text{Concat}(O_{\text{hist}}, O_{\text{future}})$
\State $z \gets E(O_{\text{full}})$ \Comment{Encode to get latent features}
\State $z_q \gets \text{VectorQuantize}(z, \mathcal{C})$ \Comment{Get discrete ground-truth tokens}
\State $z_{q,\text{hist}}, z_{q,\text{future}} \gets \text{Split}(z_q)$ \Comment{Split into history and future tokens}

\State \Comment{2. Apply masking strategy to future tokens}
\State $M \gets \text{GenerateRandomMask}(z_{q,\text{future}})$ \Comment{Create a boolean mask for future tokens}
\State $z'_{q,\text{future}} \gets \text{ApplyMaskingStrategy}(z_{q,\text{future}}, M)$ \Comment{Replace, mask, or keep tokens}

\State \Comment{3. Predict masked tokens}
\State $z'_{q,\text{input}} \gets \text{Concat}(z_{q,\text{hist}}, z'_{q,\text{future}})$ \Comment{Combine history and masked future}
\State $\text{logits} \gets M_{\text{START}}(z'_{q,\text{input}})$ \Comment{Predict logits for the entire sequence}

\State \Comment{4. Calculate loss on masked positions}
\State $L_{\text{ce}} \gets \text{CrossEntropyLoss}(\text{logits}[M], z_{q,\text{future}}[M])$ \Comment{Loss only on masked future tokens}
\State \Return $L_{\text{ce}}$
\end{algorithmic}
\end{algorithm}

Alg.~\ref{alg:inference_prediction} outlines the inference procedure. The encoded features of the historical frames serve as conditional input, while future frames are initialized with mask tokens. The MaskSTART module synthesizes future frames iteratively. In each iteration, the model predicts the probability distribution over the codebook for all masked positions, samples tokens for high-confidence positions, and re-masks low-confidence tokens to refine subsequent generation steps. This iterative refinement process continues until the entire token sequence is generated. To enhance the quality of the output, a mask scheduling strategy is employed. During the early stages of inference, the generation of the most frequent tokens (often corresponding to background classes, such as ground or sky) is suppressed, encouraging the model to prioritize the generation of key scene elements like vehicles and pedestrians. Additionally, classifier-free guidance (CFG) is employed to balance diversity and conditional accuracy during generation. This is formulated as:
\begin{equation}
    z' = z^c + \gamma \times (z^c - z^u)
\end{equation}
where $z^c$ is the conditionally generated output, $z^u$ is the unconditionally generated output (with historical conditions removed), and $\gamma$ is the guidance coefficient controlling the trade-off between consistency and diversity.

\begin{algorithm}[t]
\caption{Iterative Inference for Prediction with MaskSTART}
\label{alg:inference_prediction}
\begin{algorithmic}[1]
\Input Historical observation sequence $O_{\text{hist}}$, number of generation steps $N_{\text{iter}}$, guidance scale $\gamma$.
\Output Predicted future token sequence $\hat{z}_{q,\text{future}}$.

\State \Comment{1. Initialize with historical context and masked future}
\State $z_{q,\text{hist}} \gets \text{VectorQuantize}(E(O_{\text{hist}}))$
\State $\hat{z}_{q,\text{future}} \gets \text{InitializeWithMasks}(\text{length}=T-t)$ \Comment{Create a sequence of [MASK] tokens}

\For{$k = 1 \to N_{\text{iter}}$}
    \State \Comment{2. Predict logits with Classifier-Free Guidance}
    \State $c \gets z_{q,\text{hist}}$ \Comment{Conditional context}
    \State $u \gets \text{null\_context}$ \Comment{Unconditional context (e.g., empty sequence)}
    \State $\text{logits}_c \gets M_{\text{START}}(\text{Concat}(c, \hat{z}_{q,\text{future}}))$
    \State $\text{logits}_u \gets M_{\text{START}}(\text{Concat}(u, \hat{z}_{q,\text{future}}))$
    \State $\text{logits} \gets \text{logits}_u + \gamma \times (\text{logits}_c - \text{logits}_u)$ \Comment{Apply CFG}

    \State \Comment{3. Apply mask scheduling}
    \State $\text{logits} \gets \text{ApplyMaskScheduling}(\text{logits}, k)$ \Comment{Suppress background tokens in early steps}
    \State $\text{probs} \gets \text{Softmax}(\text{logits})$
    
    \State \Comment{4. Sample high-confidence tokens and re-mask others}
    \State $\text{confidences} \gets \max(\text{probs}, \text{dim}=-1)$
    \State $\text{mask\_to\_keep} \gets \text{GetHighConfidenceMask}(\text{confidences}, k)$
    \State $\text{new\_tokens} \gets \text{Sample}(\text{probs})$
    \State $\hat{z}_{q,\text{future}}[\text{mask\_to\_keep}] \gets \text{new\_tokens}[\text{mask\_to\_keep}]$ \Comment{Update confident tokens}
\EndFor

\State \Return $\hat{z}_{q,\text{future}}$
\end{algorithmic}
\end{algorithm}

\subsubsection{Generation Task}
For generation tasks, which is summarized in Alg.~\ref{alg:generation_task_4d}, the model generates new point cloud sequences conditioned on scene layouts ${\{P^1,..., P^T\}}$, without relying on historical point clouds. Each layout encodes the structural composition of the scene, providing a semantic blueprint for synthesis. To maximize the utility of layout information, the scene layout is represented explicitly in a 3D voxel space, as opposed to common top-down 2D representations. This 3D representation preserves height information, improving the model's ability to control the vertical positioning of objects, such as road signs and traffic lights, capabilities that are inherently limited in 2D layouts. After the layout is converted to spherical coordinates, its features are extracted using an N-layer adapter network. These layout features are then fused with voxelized point cloud features via element-wise addition using a zero convolution layer, ensuring that the layout information is seamlessly integrated without disrupting the backbone network during early training stages.

The fused features are passed to the MaskSTART module for conditional generation. Similar to the prediction task, masked tokens in the sequence are iteratively refined through scheduled decoding until the final point cloud sequence is produced. By leveraging the explicit 3D layout structure, the model generates highly realistic point cloud sequences that preserve geometric fidelity, temporal coherence, and semantic consistency.

The MaskSTART Module offers a unified framework for prediction and generation tasks, leveraging the discrete latent space and masked generative modeling to produce accurate and diverse LiDAR sequence outputs. Its ability to condition on historical observations or external layout information enables it to adapt to various real-world scenarios, including long-sequence prediction and controlled scene generation.

\begin{algorithm}[t]
\caption{Conditional Generation from 4D Layouts}
\label{alg:generation_task_4d}
\begin{algorithmic}[1]
\Input 4D scene layout sequence $L_{4D} = \{L^1, ..., L^T\}$.
\Input Number of iterative generation steps $N_{\text{iter}}$.
\Input Models: Adapter, MaskSTART, Decoder $D$, RenderModule.
\Output Generated point cloud sequence $\tilde{\mathcal{P}}$.

\Statex
\State \Comment{\textbf{1. Process 4D Conditional Layout}}
\State $V_{\text{layouts}} \gets \text{Voxelize3D}(L_{4D})$ \Comment{Voxelize the sequence of 3D layouts}
\State $V_{\text{layouts\_hcs}} \gets \text{ConvertToHCS}(V_{\text{layouts}})$ \Comment{Convert the 4D volume to HCS}
\State $F_{\text{layouts}} \gets \text{AdapterNetwork}(V_{\text{layouts\_hcs}})$ \Comment{Extract 4D layout features}

\Statex
\State \Comment{\textbf{2. Initialize Generation}}
\State $\hat{z}_q \gets \text{InitializeWithMasks}(\text{length}=T)$ \Comment{Create a 4D canvas of [MASK] tokens}

\Statex
\State \Comment{\textbf{3. Iterative Refinement using MaskSTART}}
\For{$k = 1 \to N_{\text{iter}}$}
    \State \Comment{Fuse 4D layout features with current token embeddings}
    \State $E_{\text{tokens}} \gets \text{GetEmbeddings}(\hat{z}_q)$ \Comment{Get embeddings for the current token sequence}
    \State $E_{\text{fused}} \gets E_{\text{tokens}} + \text{ZeroConv}(F_{\text{layouts}})$ \Comment{Element-wise fusion across the 4D volume}
    
    \State \Comment{Predict logits for masked positions}
    \State $\text{logits} \gets M_{\text{START}}(E_{\text{fused}})$
    
    \State \Comment{Apply scheduled decoding}
    \State $\text{probs} \gets \text{Softmax}(\text{ApplyMaskScheduling}(\text{logits}, k))$
    
    \State \Comment{Sample high-confidence tokens and re-mask others}
    \State $\text{confidences} \gets \max(\text{probs}, \text{dim}=-1)$
    \State $\text{mask\_to\_keep} \gets \text{GetHighConfidenceMask}(\text{confidences}, k)$
    \State $\text{new\_tokens} \gets \text{Sample}(\text{probs})$
    \State $\hat{z}_q[\text{mask\_to\_keep}] \gets \text{new\_tokens}[\text{mask\_to\_keep}]$ \Comment{Update confident tokens}
\EndFor

\Statex
\State \Comment{\textbf{4. Decode Final Token Sequence}}
\State $V_{\text{gen}} \gets D(\hat{z}_q)$ \Comment{Decode the completed 4D token sequence}
\State $\tilde{\mathcal{P}} \gets \text{RenderModule}(V_{\text{gen}})$ \Comment{Render the final point cloud sequence}

\State \Return $\tilde{\mathcal{P}}$
\end{algorithmic}
\end{algorithm}

\section{More Experiments}
\subsection{Qualitative Results}

\begin{figure*}
    \centering
    \includegraphics[width=1.0\linewidth]{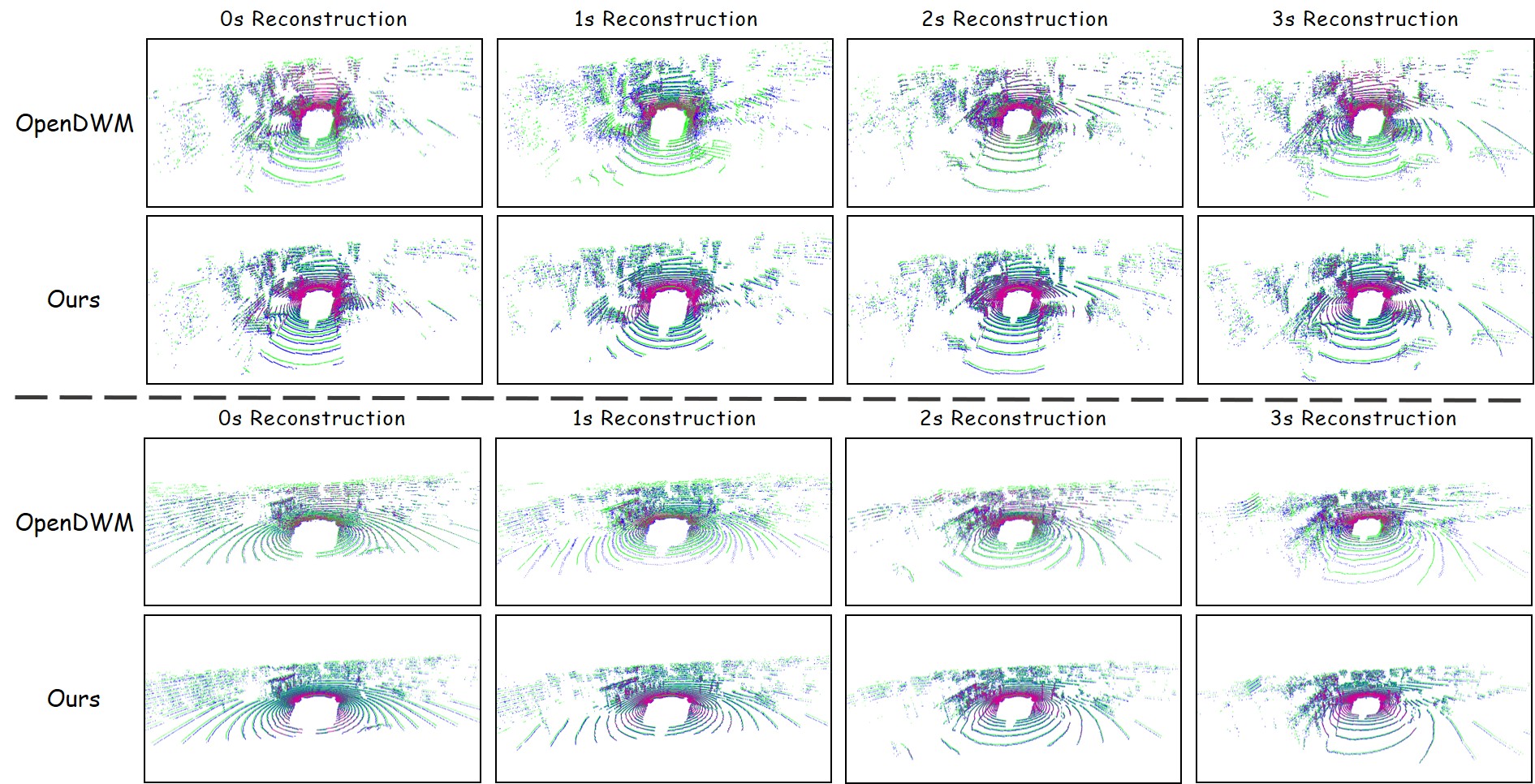}
    \caption{Qualitative comparison of point cloud reconstruction. We compare our method against the OpenDWM baseline on two distinct sequences (top and bottom sections) for time horizons of 0s, 1s, 2s, and 3s. The visualization overlays reconstructions with the ground truth: magenta indicates the correct intersection (true positives), green denotes missed ground truth (false negatives), and blue highlights reconstruction artifacts (false positives). Our method consistently produces more complete reconstructions (denser magenta) and significantly fewer artifacts (less blue) across all time steps, demonstrating superior reconstruction accuracy and robustness.}
    \label{fig:rec1}
\end{figure*}

Fig. \ref{fig:rec1} presents a qualitative comparison of our method against the OpenDWM baseline for point cloud reconstruction across two distinct sequences. The visualization overlays reconstructions with the ground truth, where magenta indicates the correct intersection (true positives), green denotes missed ground truth (false negatives), and blue highlights reconstruction artifacts (false positives). The results visually underscore the superior performance of our approach. The outputs from OpenDWM become progressively noisy and incomplete over time, accumulating significant false positives (blue artifacts) while failing to capture the full geometry. In stark contrast, our method consistently produces more complete reconstructions, evidenced by a denser volume of true positives (magenta), and maintains this high fidelity across all time steps from 0s to 3s. This demonstrates our model's enhanced ability to robustly integrate temporal information without the significant error accumulation that plagues the baseline, validating its superior accuracy and robustness.

\begin{figure*}
    \centering
    \includegraphics[width=1.0\linewidth]{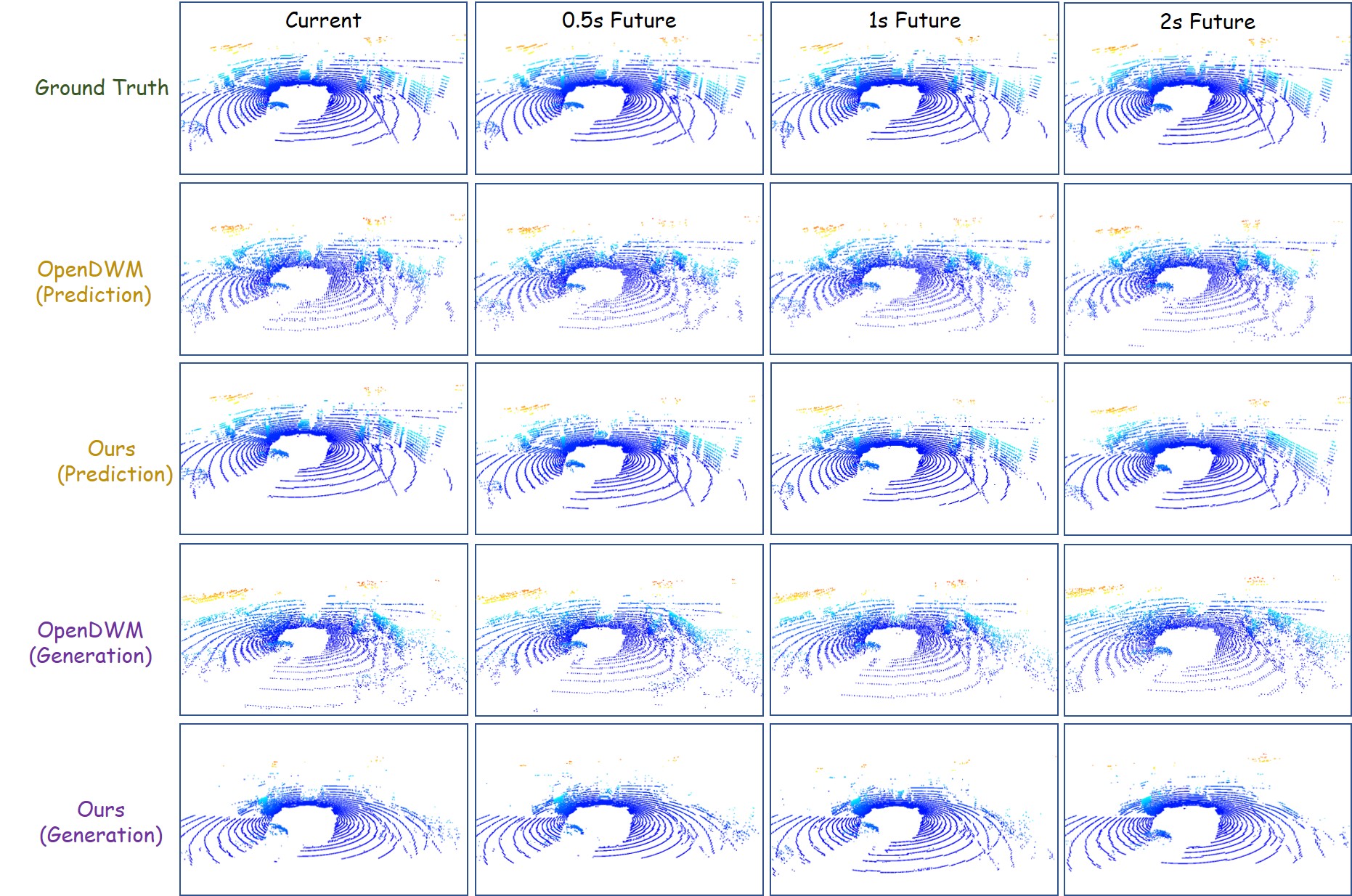}
    \caption{Qualitative results for prediction and generation. We compare our method with OpenDWM against the ground truth for future horizons up to 2s. Our method consistently produces sharper and more accurate results for both static background and dynamic objects (highlighted) compared to the baseline. The baseline's predictions and generations degrade significantly over time, losing structural detail.}
    \label{fig:pre1}
\end{figure*}

Fig. \ref{fig:pre1} provides a qualitative comparison of our method against the OpenDWM baseline for future LiDAR prediction and generation. The results visually underscore the superior fidelity and noise handling of our approach. The outputs from OpenDWM suffer from significant noise, particularly in the far-field, where it hallucinates numerous scattered points. Furthermore, its representation of near-field objects becomes progressively blurry and loses structural integrity over time. In stark contrast, our method generates much cleaner sequences, effectively suppressing far-field noise while maintaining a dense and geometrically accurate representation of near-field objects. As highlighted in the figure, our model consistently produces sharp, well-defined structures that closely match the ground truth, demonstrating a superior ability to model complex 4D dynamics with both high fidelity and robustness to noise.

\subsection{Limitations}
Despite the strong performance of LiSTAR, we acknowledge several limitations that present opportunities for future work. First, our HCS representation is specifically tailored to the geometry of spinning LiDARs. This specialization, while effective, may limit its direct applicability to other 3D sensor modalities, such as solid-state LiDARs or depth cameras, which feature different sampling patterns. Second, as a VQ-VAE-based model, LiSTAR is subject to inherent quantization error, where fine-grained details can be lost during the discretization of the latent space. Furthermore, the iterative refinement process of the MaskSTART module, while crucial for high-quality synthesis, incurs higher computational latency during inference compared to single-pass generative models, which could be a consideration for real-time applications. Finally, our controllable generation relies on the availability of detailed 4D point cloud-aligned voxel layouts, which may not always be accessible in all scenarios. Future research could focus on developing more universal representations, exploring faster generative paradigms, and enabling more abstract forms of conditioning, such as natural language commands.

\end{document}